\documentclass{article}

 \usepackage[preprint]{neurips_2026}

\usepackage[utf8]{inputenc} 
\usepackage[T1]{fontenc}    
\usepackage{hyperref}       
\usepackage{url}            
\usepackage{booktabs}       
\usepackage{amsfonts}       
\usepackage{nicefrac}       
\usepackage{microtype}      
\usepackage[table]{xcolor}
\definecolor{softgray}{RGB}{242,244,246}

\usepackage{amsmath, amsthm}
\usepackage{graphicx}
\usepackage{multirow}
\usepackage{amssymb}
\usepackage{pifont}
\usepackage{wrapfig}
\usepackage{subcaption}
\usepackage{adjustbox}
\usepackage{placeins}

\usepackage[normalem]{ulem}
\useunder{\uline}{\ul}{}

\usepackage{float}
\usepackage{algorithm}
\usepackage{algpseudocode}

\title{ConRetroBert: EMA Stabilized Dual Encoders for Template-Based Single-Step Retrosynthesis}

\author{%
  Mohammad Jahid Ibna Basher \\
  Department of Industrial Engineering\\
  University of Central Florida\\
  Orlando, FL 32816 \\
  \texttt{mo253203@ucf.edu} \\
  \And
  Ali Khodabandeh Yalabadi \\
  Department of Industrial Engineering\\
  University of Central Florida\\
  Orlando, FL 32816 \\
  \texttt{yalabadi@ucf.edu} \\
  \AND
  Ivan Garibay \\
  Department of Industrial Engineering\\
  University of Central Florida\\
  Orlando, FL 32816 \\
  \texttt{igaribay@ucf.edu} \\
  \And
  Ozlem Ozmen Garibay \\
  Department of Industrial Engineering\\
  University of Central Florida\\
  Orlando, FL 32816 \\
  \texttt{ozlem@ucf.edu} \\
}

\begin{document}

\maketitle

\begin{abstract}
Template-based single step retrosynthesis predicts reactants by selecting and applying an explicit reaction template, making each prediction traceable to a chemical transformation rule. This interpretability is useful for synthesis planning, but template-based methods are often viewed as less competitive than template free models because template prediction is commonly formulated as global classification over a long tailed rule library. We argue that this weakness is not inherent to templates, but to the learning formulation. We present \textbf{ConRetroBert}, a dual encoder framework that reframes template-based retrosynthesis as dense product template retrieval followed by candidate set listwise ranking. In Stage 1, contrastive pretraining learns a shared embedding space between products and reaction templates. In Stage 2, a multi positive listwise objective refines template ranking over mined hard negative candidate sets, matching the inference time decision problem more closely than full vocabulary classification. To enable template side adaptation without destabilizing hard negative mining, ConRetroBert uses a slow moving exponential moving average, or EMA, template encoder for retrieval bank construction while updating the live template encoder through the ranking loss. On the local USPTO-50k benchmark, the main gain comes from Stage 2 candidate set ranking, which improves top-1 reaction accuracy from $50.5\%$ to about $61.3\%$, while EMA stabilized template adaptation provides a further improvement to about $62.4\%$. The local model reaches $81.6\%$, $85.3\%$, and $87.8\%$ at top-3, top-5, and top-10. We further show that retrieval based template prediction is especially strong in the long tail of rare templates, and that Stage 2 ranking improves the applicability of retrieved templates relative to contrastive retrieval alone. As a separate scaling result, fine tuning from a leakage controlled USPTO-Full checkpoint reaches $75.4\%$ top-1 accuracy on USPTO-50k. These results show that template-based retrosynthesis can combine strong predictive performance with chemically inspectable predictions, challenging the common assumption that high accuracy requires abandoning explicit reaction templates. The code and data is provided at this \href{https://github.com/JahidBasher/ConRetroBert}{GitHub URL}.
\end{abstract}

\section{Introduction}
Computer aided retrosynthesis aims to recover plausible reactant sets for a target product molecule and is a core component of modern synthesis planning systems \citep{coley2017computer,segler2017neural}. In practical multistep synthesis planning, a single step retrosynthesis model is repeatedly queried to expand search nodes, and errors in early disconnection decisions can propagate through the entire route construction process \citep{torren2024models}. Improving single step retrosynthesis is therefore not only a matter of benchmark accuracy. A useful model should also provide predictions that are reliable, inspectable, and compatible with downstream planning constraints.

Recent progress has been driven largely by template-free and semi-template methods, including sequence generation, graph editing, multitask graph learning, and flow based generative models \citep{zheng2019predicting,wan2022retroformer,sacha2021molecule,zhong2023retrosynthesis,chen2023g,wang2023retrosynthesis,zhao2025single,yadav2025retro}. These methods are attractive because they are not explicitly restricted to a fixed reaction template library, and several recent systems report strong benchmark performance on USPTO-50k. In particular, RetroSynFlow reports strong top-$k$ accuracy and presents discrete flow matching as a flexible route toward accurate and diverse single step retrosynthesis \citep{yadav2025retro}. At the same time, recent work on interpretable and structured retrosynthesis highlights a persistent limitation of many high performing systems: a model may generate a plausible reactant set while providing limited mechanistic evidence for why a particular disconnection was selected \citep{wang2023retrosynthesis,zhao2025single}.

Template-based retrosynthesis offers a complementary set of strengths. Instead of generating reactants directly, template-based systems predict an explicit reaction rule, usually encoded as a SMARTS transformation, and apply that rule to the product molecule \citep{coley2017computer,segler2017neural,dai2019retrosynthesis}. This makes the prediction traceable by construction. Each output can be associated with a concrete transformation pattern that can be inspected, validated by a reaction engine, filtered by chemical constraints, or used as an action inside a planner. These properties are especially valuable in scientific and planning settings, where users often need to understand not only the predicted reactants, but also the chemical operation that produced them.

Despite this advantage, template-based methods are often viewed as less scalable and less competitive than template-free models. We argue that this limitation is not inherent to templates themselves, but to the dominant learning formulation used for template prediction. Many template-based systems cast the problem as global classification over a long tailed library of reaction templates \citep{dai2019retrosynthesis,chen2021deep}. This creates three coupled issues. First, frequent templates dominate the loss, making rare but useful transformations difficult to learn. Second, the softmax denominator treats every unobserved template as a competing label, even though several unrecorded templates may be chemically applicable or may lead to plausible alternative disconnections for the same product. This label incompleteness is not eliminated by our method, but candidate-set ranking reduces the problem from full library competition to the chemically similar alternatives that determine the inference time decision. Third, the output layer grows with the template library, making global classification increasingly unattractive as template sets scale toward hundreds of thousands of transformations.

In this work, we revisit template-based retrosynthesis from a retrieval learning perspective. We present \textbf{ConRetroBert}, a dual encoder framework for template-based single step retrosynthesis that treats reaction templates as searchable chemical rules rather than as class labels in a large output vocabulary. A product encoder and a template encoder map products and reaction templates into a shared embedding space, so template prediction becomes nearest neighbor retrieval followed by ranked template application.

ConRetroBert is most closely related to retrieval-oriented template-based retrosynthesis methods and to dense retrieval training with hard negatives and slow-moving encoder targets; a fuller discussion is given in Appendix~\ref{app:related_work_full}. We do not claim retrieval itself as the novelty: prior methods such as MHNreact and local template retrieval already use product-template embedding or neighborhood-based evidence. Our contribution is to combine dual-encoder contrastive product-template pretraining with candidate-set hard-negative listwise ranking and EMA-stabilized template-side adaptation in a single template-based retrosynthesis framework.

ConRetroBert follows a two stage training strategy. Stage 1 learns a joint product template representation space using symmetric contrastive pretraining, aligning each product with its observed reaction template while using other examples in the batch as negatives. Stage 2 then refines this representation into a template proposal policy through multi positive listwise ranking over dynamically mined hard negative candidate sets. This second stage is the central step in the method. It replaces full vocabulary competition with candidate-set ranking, so the model learns to separate observed positive templates from the chemically and geometrically similar alternatives that are actually retrieved at test time. In our experiments, this change is also the main source of the performance gain: moving from Stage 1 to the frozen template encoder Stage 2 model improves top-1 reaction accuracy by $10.9$ percentage points in the local USPTO-50k regime.

A central technical challenge arises when both encoders in the dual-encoder architecture are allowed to adapt. The product encoder acts as the query model and should update rapidly, whereas the template encoder defines the embedding space used to construct the hard-negative retrieval index. If the template encoder changes too quickly, the index becomes stale, and the mined negatives may no longer reflect the model’s current scoring function. To address this issue, ConRetroBert introduces an EMA-stabilized trainable template encoder. The live template encoder is updated by gradients from the listwise ranking objective, while a slow-moving EMA shadow encoder is used to rebuild the retrieval index at the epoch level. EMA therefore serves as a stabilization mechanism that enables template-side adaptation without disrupting candidate construction. It should not be interpreted as the primary source of the method’s gains; rather, the principal improvement comes from the Stage 2 candidate-set ranking formulation.

Empirically, the local USPTO-50k regime shows that Stage 2 EMA improves mean top-1 reaction accuracy from $61.28\%$ to $62.44\%$ relative to Frozen Template Encoder (TE), while remaining stable across repeated runs. We also find that retrieval based template prediction is especially strong in the long tail, and that Stage 2 improves the applicability of retrieved templates relative to Stage 1. As a separate scaling result, fine tuning from a leakage controlled USPTO-Full checkpoint reaches 75.4\% top-1 accuracy. These results show that explicit reaction templates can remain competitive while preserving rule level transparency for chemical interpretation and synthesis planning.

Our contributions are as follows. First, we formulate template-based single step retrosynthesis as dense product template retrieval followed by candidate set listwise ranking, replacing global template classification with an objective that better matches inference time decisions. Second, we introduce an EMA stabilized trainable template encoder that enables controlled template side adaptation while maintaining a stable retrieval index for hard-negative mining. Third, we provide a careful evaluation pipeline for template-based prediction, including reactant-set ranking, template application, deduplication, and leakage controlled transfer from USPTO-Full to USPTO-50k. Fourth, we show through local, long tail, reliability, and matched budget transfer analyses that candidate-set ranking substantially strengthens template-based retrosynthesis while retaining explicit reaction rule traceability.


\vspace{-3mm}
\section{Methodology}
\label{method}

ConRetroBert reframes template-based single step retrosynthesis as retrieval and ranking over explicit chemical transformation rules. Given a product molecule, the model retrieves candidate reaction templates from a learned embedding space, ranks them with a listwise objective, and applies the highest ranked templates to generate reactant sets. This section describes the core model. A schematic overview of the full training and inference pipeline is provided in Appendix~\ref{app:method_details}, Figure~\ref{fig:conretro_pipeline}. Additional architectural, algorithmic, mathematical, curation, and evaluation details are provided in Appendices \ref{app:method_details}--\ref{app:data_curation}.

\subsection{Problem Formulation}
\label{sec:problem_formulation}

Let $x$ denote a product molecule and let $\mathcal{T}=\{\tau_1,\ldots,\tau_N\}$ be a template library containing $N$ reaction templates. Each template $\tau$ is represented as a SMARTS transformation and can be applied to $x$ through a reaction engine to produce zero or more candidate reactant sets. For each product $x$, let $P(x)\subseteq \mathcal{T}$ denote the set of observed positive templates associated with the product in the training data. This set can contain more than one template because the same product may appear in multiple reactions or be consistent with multiple recorded transformations.

We learn a scoring function
\begin{equation}
s_\theta(x,\tau)
=
\left\langle z_P(x;\theta_P), z_T(\tau;\theta_T) \right\rangle,
\end{equation}
where $z_P$ and $z_T$ are product and template embeddings produced by two encoder towers. At inference, templates are ranked by $s_\theta(x,\tau)$, the top templates are applied to $x$, and the resulting reactant sets are canonicalized, deduplicated, and ranked by their template scores.

\subsection{Dual Encoder Template Retrieval}
\label{sec:dual_encoder}

ConRetroBert uses two independent Transformer encoders with the same character level vocabulary but separate parameters. One tower encodes product SMILES strings and the other encodes template SMARTS strings. Each tower consists of token embeddings, learnable positional embeddings, and $L$ Transformer encoder layers. We use the first token hidden state directly as the sequence embedding:
\begin{equation}
z_P(x)=f_P(x;\theta_P),
\qquad
z_T(\tau)=f_T(\tau;\theta_T).
\end{equation}
Both embeddings are $\ell_2$ normalized, and the final score is temperature scaled cosine similarity:
\begin{equation}
s_\theta(x,\tau)
=
\frac{\hat{z}_P(x)^\top \hat{z}_T(\tau)}
{\tau_{\mathrm{temp}}}.
\end{equation}

This architecture replaces a global template classifier with a searchable template library. The output layer no longer grows with the number of templates, and training can focus on candidate level ranking rather than full library classification.

\subsection{Stage 1: Contrastive Product-Template Alignment}
\label{sec:stage1}

Stage 1 learns the shared product template retrieval space. Given a minibatch of $B$ paired examples $\{(x_i,\tau_i)\}_{i=1}^B$, we compute the similarity matrix $S\in\mathbb{R}^{B\times B}$ with entries $S_{ij}=s_\theta(x_i,\tau_j)$. We train both encoders using a symmetric contrastive loss:
\begin{equation}
\mathcal{L}_{\mathrm{S1}}
=
\frac{1}{2}
\left[
\frac{1}{B}\sum_{i=1}^{B}
-\log
\frac{\exp(S_{ii})}
{\sum_{j=1}^{B}\exp(S_{ij})}
+
\frac{1}{B}\sum_{j=1}^{B}
-\log
\frac{\exp(S_{jj})}
{\sum_{i=1}^{B}\exp(S_{ij})}
\right].
\label{eq:stage1}
\end{equation}
The first term aligns each product to its paired template, while the second aligns each template to its paired product. Other examples in the batch serve as negatives. After Stage 1, all template embeddings can be precomputed and searched efficiently.

\subsection{Stage 2: Multi Positive Listwise Ranking}
\label{sec:stage2}

Stage 2 trains a multi-positive listwise ranking objective for reaction templates given a product query. Instead of assuming a single correct template per product, it treats all templates observed with the same product SMILES in the training split as observed positives. These positives are obtained by grouping reaction records by product and taking the union of their associated templates, so no additional annotation is required. However, they are only observationally positive: they reflect templates recorded in the dataset, not the full set of chemically applicable templates. Consequently, some mined hard negatives may be chemically valid but unobserved alternatives, so Stage 2 should be interpreted as a ranking objective under incomplete positive labels rather than as a complete enumeration of all valid disconnections.

Stage 1 creates a useful retrieval space, but contrastive pretraining alone does not directly optimize the inference time ranking problem. Stage 2 therefore trains a template proposal policy over a candidate set $\mathcal{C}(x)$ assembled for each product. This candidate set always contains all known positives $P(x)$ together with hard negatives and in-batch negatives. Hard negatives are retrieved from a FAISS index over template embeddings.

For a product $x$, the policy over $\mathcal{C}(x)$ is
\begin{equation}
\pi_\theta(\tau\mid x)
=
\frac{\exp(s_\theta(x,\tau))}
{\sum_{\tau'\in \mathcal{C}(x)}\exp(s_\theta(x,\tau'))}.
\end{equation}
Because multiple templates may be valid positives for the same product, we optimize a multi-positive listwise negative log likelihood:
\begin{equation}
\mathcal{L}_{\mathrm{rank}}(x)
=
-\frac{1}{|P(x)|}
\sum_{\tau\in P(x)}
\log \pi_\theta(\tau\mid x).
\label{eq:stage2_rank}
\end{equation}
We additionally use label smoothing and a small entropy term to discourage overconfident collapse:
\begin{equation}
\mathcal{L}_{\mathrm{S2}}(x)
=
\mathcal{L}_{\mathrm{rank}}(x)
-
\beta H(\pi_\theta(\cdot\mid x)).
\label{eq:stage2_full}
\end{equation}

where $H(\pi_\theta(\cdot\mid x))$ denotes the entropy of the candidate-set policy.

For readability, Eq.~\ref{eq:stage2_full} shows the unsmoothed form; the implementation uses the label-smoothed target distribution given in Appendix~\ref{app:label_smoothing}. The key distinction from global classification is that the model learns from the templates most likely to be confused with the correct rule, rather than from a dense denominator dominated by easy negatives. A unified view of all reported Stage 2 variants and additional mathematical justification for this objective are provided in Appendix~\ref{app:stage2_algorithm} and Appendix~\ref{app:method_math}.

\subsection{EMA-Stabilized Template Encoder}
\label{sec:ema_encoder}

A frozen template encoder provides a stable retrieval index, but it prevents template representations from adapting to the Stage 2 ranking objective. Updating both encoders naively introduces a different problem: the template embeddings define the index used for hard-negative mining, so rapid template encoder updates can make the retrieval index stale within an epoch. ConRetroBert addresses this with a two timescale template encoder.

We maintain a live template encoder with parameters $\theta_T$ and an EMA shadow encoder with parameters $\bar{\theta}_T$. The live encoder receives gradient updates from the listwise ranking objective, while the EMA shadow is updated without gradients:
\begin{equation}
\bar{\theta}_T
\leftarrow
\alpha \bar{\theta}_T
+
(1-\alpha)\theta_T,
\qquad
\alpha=0.999.
\label{eq:ema_update}
\end{equation}
At the beginning of each epoch, the full template library is re encoded using the EMA shadow:
\begin{equation}
Z_T^{\mathrm{ema}} = \{ z_T(\tau;\bar{\theta}_T) : \tau \in \mathcal{T} \}.
\end{equation}
This cache is used to rebuild the retrieval index and mine hard negatives for the epoch. The live template encoder can therefore adapt to the ranking loss, while the hard negative index evolves smoothly enough to remain stable during training.

We optimize the product and template towers with separate optimizers. The product encoder uses a larger learning rate because it acts as the query model, while the template encoder uses a smaller learning rate because it defines the retrieval basis. This separation is important empirically: one optimizer co training without EMA destabilizes Stage 2 and underperforms even the Stage 1 model. Full optimizer settings and ablations are reported in Appendix~\ref{app:method_details}.
Appendix~\ref{app:method_math} gives a short stability analysis showing how the EMA teacher smooths the motion of the template retrieval bank relative to the live template encoder.

\subsection{Template Application and Reactant-Set Ranking}
\label{sec:template_application}

The model outputs ranked templates, but benchmark accuracy is computed over ranked reactant sets. We therefore convert template rankings into reactant-set rankings before evaluation. For each product, we retrieve the top templates, apply each template to the product, collect up to a fixed number of outcomes per template, canonicalize each reactant set with RDKit, and deduplicate identical canonical reactant keys. If multiple templates generate the same reactant set, we keep the highest template score. The resulting unique reactant sets are sorted by score and used for top $k$ evaluation. Rows for which all retrieved templates are multi-input and therefore yield no prediction are counted as misses in the reported accuracy denominators. The full evaluation protocol is given in Appendix~\ref{app:evaluation_protocol}.

\subsection{Data Curation and Leakage Control}
\label{sec:data_curation_short}

We use USPTO-50k as the primary benchmark and USPTO-Full for large-scale transfer initialization. Because USPTO-50k is derived from the broader USPTO corpus, the transfer setting requires explicit leakage control. Therefore, before pretraining or transfer initialization, we remove all USPTO-50k validation and test reactions from the USPTO-Full training corpus using canonical reaction signatures. We also exclude invalid templates and multi-input templates that cannot be applied consistently by the reaction engine. For USPTO-50k itself, we retain the standard benchmark split without additional filtering to remain comparable with prior work. Details of the curation procedure and dataset statistics are provided in Appendix~\ref{app:data_curation}.
\vspace{-3mm}
\section{Experiments}
\label{exp}

We evaluate ConRetroBert on USPTO-50k in a local regime and a transfer regime from USPTO-Full. The local regime is the cleanest test of the proposed ranking formulation, while the transfer regime is reported as a scaling result. Additional implementation details and extended results are provided in Appendices \ref{app:exp_details} and \ref{app:extended_results}.

\subsection{Experimental Setup}
\label{sec:exp_setup}

Our primary benchmark is USPTO-50k under the standard GLN split \citep{dai2019retrosynthesis}. All local experiments are trained on this corpus only. For transfer experiments, we pretrain on a leakage-cleaned USPTO-Full corpus and then fine-tune on USPTO-50k. Data curation details are provided in Appendix~\ref{app:data_curation}.

We evaluate at the reactant set level. For each product, we retrieve top-ranked templates, apply each template to the product, canonicalize and deduplicate the resulting reactant sets, and rank unique reactant candidates by the highest score of any template that generated them. Multi-input template failures that yield no prediction are counted as misses in the accuracy denominator. The exact evaluation protocol is given in Appendix~\ref{app:evaluation_protocol}.

We report \emph{reaction top-$k$ accuracy} after template application and reactant-set deduplication, \emph{template retrieval top-$k$ accuracy} before template application, and diagnostic measures of template applicability and unique reactant sets to assess whether retrieved templates remain chemically usable actions.

\subsection{Comparison with Prior Work}
\label{sec:exp_sota}

We first compare the best \emph{local} ConRetroBert model against representative published methods on USPTO-50k. We separate this comparison from transfer results to avoid conflating local method validation with gains from larger-scale pretraining. Since prior methods may differ in curation, preprocessing, and training data conditions, we present these comparisons as reported-number context rather than as claims of strict protocol equivalence.
Among our local variants, \textbf{Stage 2 EMA Mid} denotes the EMA stabilized Stage 2 model with three trainable template encoder layers; full variant definitions are given in Appendix~\ref{app:variant_definitions}.

\begin{table}[t]
\centering
\caption{Top-k reaction accuracy (\%) on USPTO-50k. TB = template-based, ST = structured or semi-template, TF = template-free. ConRetroBert rows use the local USPTO-50k regime; EMA Mid values are rounded from the three-run mean in Table~\ref{tab:local_main}. Prior numbers are taken from the RetroSynFlow comparison table and may differ in preprocessing. Best local result in each column is in \textbf{bold}.}
\label{tab:sota_local}
\small
\setlength{\tabcolsep}{4.2pt}
\renewcommand{\arraystretch}{1.12}
\begin{tabular}{@{}llcccc@{}}
\toprule
 & \textbf{Model} & $k{=}1$ & $k{=}3$ & $k{=}5$ & $k{=}10$ \\
\midrule
\multirow{4}{*}{TB}
& GLN~\cite{dai2019retrosynthesis}           & 52.5 & 74.7 & 81.2 & 87.9 \\
& LocalRetro~\cite{chen2021deep}             & 52.6 & 76.0 & 84.4 & 90.6 \\
& RetroGFN~\cite{gainski2025diverse}         & 49.2 & 73.3 & 81.1 & 88.0 \\
\midrule
\multirow{5}{*}{ST / TF}
& GraphRetro~\cite{somnath2021learning}      & 53.7 & 68.3 & 72.2 & 75.5 \\
& MEGAN~\cite{sacha2021molecule}             & 48.0 & 70.9 & 78.1 & 85.4 \\
& Retroformer~\cite{wan2022retroformer}      & 52.9 & 68.2 & 72.5 & 76.4 \\
& Chimera~\cite{maziarz2024chimera}          & 59.6 & 82.8 & 89.2 & 94.2 \\
& RetroSynFlow~\cite{yadav2025retro}         & 60.0 & 77.9 & 82.7 & 85.3 \\
\midrule
\multicolumn{6}{@{}l}{\textit{Ours, local USPTO-50k regime}} \\
& ConRetroBert Stage 1                       & 50.5 & 72.5 & 78.8 & 83.5 \\
& ConRetroBert Stage 2 Frozen-TE             & 61.4 & 78.0 & 82.3 & 85.7 \\
\rowcolor{softgray}
& \textbf{ConRetroBert Stage 2 EMA Mid}      & \textbf{62.5} & \textbf{81.6} & \textbf{85.3} & \textbf{87.8} \\
\bottomrule
\end{tabular}
\vspace{-5mm}
\end{table}

Under our reported local USPTO-50k setting, ConRetroBert Stage 2 EMA Mid achieves the strongest reported top-1 result among the methods listed in Table~\ref{tab:sota_local}, including RetroSynFlow at $60.0\%$ \citep{yadav2025retro}. At larger cutoffs, however, several published baselines remain stronger, including Chimera and LocalRetro at top-10. We therefore interpret the local comparison narrowly: the main strength of ConRetroBert is a sharper top-of-list decision, not universal domination across all k. Because MHNreact is the closest retrieval-oriented template prior, Appendix~\ref{app:mhnreact_comparison} provides a separate protocol-caveated comparison to its reported USPTO-50k numbers.

\subsection{Local Regime: Main Method Validation}
\label{sec:exp_local}

The local regime provides the cleanest validation of the proposed method. Table~\ref{tab:local_main} reports the main local progression from Stage 1 retrieval to Stage 2 ranking, together with the most important ablations.

\begin{table}[t]
\centering
\caption{Local USPTO-50k results. Reaction top-k is the main metric and template retrieval top-k is a diagnostic. Frozen-TE, EMA Mid, and Snapshot + KLD are reported as mean $\pm$ std over three independent runs. Bold indicates the best mean result in each column.}
\label{tab:local_main}
\small
\setlength{\tabcolsep}{4.0pt}
\renewcommand{\arraystretch}{1.12}
\begin{tabular}{@{}lcccccc@{}}
\toprule
\multirow{2}{*}{\textbf{Method}} &
\multicolumn{4}{c}{\textbf{Reaction Acc. (\%)}} &
\multicolumn{2}{c}{\textbf{TRetr. (\%)}} \\
\cmidrule(lr){2-5}\cmidrule(l){6-7}
& Top-1 & Top-3 & Top-5 & Top-10 & Top-1 & Top-10 \\
\midrule
Stage 1 (S1)                             & 50.5 & 72.5 & 78.8 & 83.5 & 21.4 & 59.7 \\
S2 Frozen-TE ($|\mathcal{C}|{=}64$) & 61.28$\pm0.12$ & 77.93$\pm0.09$ & 82.51$\pm0.14$ & 85.74$\pm0.06$ & \textbf{38.27}$\pm0.44$ & 67.85$\pm0.15$ \\
\rowcolor{softgray}
S2 EMA Mid ($K{=}3$)            & \textbf{62.44}$\pm0.09$ & \textbf{81.72}$\pm0.23$ & \textbf{85.29}$\pm0.12$ & \textbf{87.71}$\pm0.16$ & 37.25$\pm0.33$ & 70.23$\pm0.22$ \\
\midrule
\multicolumn{7}{@{}l}{\textit{Selected ablations}} \\
Snapshot + KLD                       & 61.98$\pm0.11$ & 81.67$\pm0.17$ & 85.19$\pm0.12$ & 87.50$\pm0.12$ & 36.81$\pm0.19$ & \textbf{70.39}$\pm0.31$ \\
EMA + ALT ($10{:}2$)                 & 60.7 & 79.5 & 84.2 & 86.9 & 35.6 & 68.0 \\
One-Opt (no refresh)                 & 49.5 & 70.1 & 75.8 & 80.5 & 20.4 & 55.7 \\
\bottomrule
\end{tabular}
\vspace{-5mm}
\end{table}

Three findings are most important. First, Stage 2 ranking is itself a major improvement over Stage 1 retrieval alone: moving from Stage 1 to the frozen template encoder Stage 2 baseline yields a gain of $+10.9$ points in top-1 reaction accuracy and $+17.3$ points in template retrieval top-1. This confirms that candidate-set ranking better matches the inference problem than contrastive alignment alone.

Second, EMA stabilized template adaptation provides a further gain over the frozen template encoder baseline. Across three independent runs, Stage 2 EMA Mid improves mean top-1 reaction accuracy from $61.28\%$ to $62.44\%$ relative to Frozen-TE, and mean top-10 reaction accuracy from $85.74\%$ to $87.71\%$. The observed variation is small for both models, with standard deviations of $0.12$ and $0.09$ at top-1 for Frozen-TE and EMA Mid respectively, supporting the view that the EMA gain is repeatable rather than a one-run artifact.

Third, not all additional mechanisms help. Snapshot + KLD is competitive but remains below EMA Mid in mean performance across the three runs, alternating freeze is weaker in the local setting, and one-optimizer co-training collapses below the Stage 1 reference point. These negative results strengthen the claim that the gain comes from stable template-side adaptation rather than simply training more parameters.

A notable asymmetry remains between template retrieval and final reactant accuracy. EMA Mid is slightly below the frozen baseline on template top-1 retrieval, but improves reaction accuracy from top-3 onward. This indicates that final retrosynthesis accuracy is not determined solely by the rank of the recorded ground-truth template. Instead, EMA improves the quality and ordering of the retrieved set in a way that better supports the final template-to-reactant conversion. Viewed together with Table~\ref{tab:sota_local}, this also suggests that the local EMA model sharpens the head of the ranking more than it expands the far tail, which helps explain why its strongest advantage appears at top-1 rather than at top-10 against all previously reported baselines.

\subsection{Auxiliary Long-Tail Template Frequency Analysis}
\label{sec:exp_longtail}

Template prediction is highly imbalanced, with a small set of frequent templates and a large tail of rare templates. As supporting evidence for the retrieval-based design choice, we compare classification and retrieval scoring on a shared backbone, grouped by the training frequency of the ground-truth template. This is an auxiliary scoring diagnostic rather than a full Stage-2 pipeline comparison.

We partition test reactions into head templates with frequency \(f>5\), tail templates with \(0<f\leq 5\), and open-library templates with \(f=0\) in the training split but present in the inference template library. The retrieval model can score these available template strings by similarity, whereas the classifier has output classes only for templates observed during training and cannot predict \(f=0\) labels by construction. Thus Table~\ref{tab:longtail_main} should be read as an open-library scoring diagnostic, not as a closed-vocabulary unseen-class comparison.

\begin{table}[t]
\centering
\caption{Auxiliary long-tail analysis by ground-truth template frequency on USPTO-50k. Values are template top-$k$ accuracy for a discriminative classifier and a contrastive retrieval model built on the same ConRetroBert backbone, rather than reactant-set accuracy of the complete ConRetroBert Stage-2 pipeline. Head: $f > 5$, Tail: $0 < f \le 5$, Open-lib: $f = 0$ in the training split but present in the inference template library. Best value in each row is in bold.}
\label{tab:longtail_main}
\small
\setlength{\tabcolsep}{4.0pt}
\renewcommand{\arraystretch}{1.12}
\begin{tabular}{@{}llccc@{}}
\toprule
\textbf{$k$} & \textbf{Model} & \textbf{Head} & \textbf{Tail} & \textbf{Open-lib} \\
\midrule
\multirow{2}{*}{Top-1}
& Classifier & \textbf{53.76} & 8.46 & 0.00 \\
& Retrieval  & 53.56 & \textbf{18.68} & \textbf{9.19} \\
\midrule
\multirow{2}{*}{Top-10}
& Classifier & \textbf{91.84} & 48.05 & 0.00 \\
& Retrieval  & 84.25 & \textbf{61.09} & \textbf{37.25} \\
\bottomrule
\end{tabular}
\vspace{-5mm}
\end{table}

The pattern supports the retrieval motivation: classification is slightly stronger on head templates, while retrieval is stronger on tail templates and can score training-unseen templates when they are available in the inference library. Additional details are reported in Appendix~\ref{app:longtail_appendix}.

\subsection{Reliability and Applicability of Retrieved Templates}
\label{sec:exp_reliability}

To test whether explicit templates provide a useful reliability layer, we measure how often retrieved templates are chemically applicable and how many distinct reactant sets they generate. Table~\ref{tab:reliability_main} compares Stage 1, Stage 2 Frozen-TE, and Stage 2 EMA Mid, while Appendix~\ref{app:case_studies} provides qualitative examples showing that the retrieved template remains chemically inspectable in both success and failure.

\begin{table}[t]
\centering
\caption{Reliability diagnostics on USPTO-50k. AppRate@$k$ is the fraction of retrieved templates that successfully apply to the product. UniqueRS@$k$ is the mean number of distinct canonicalized reactant sets generated from the top-$k$ templates. Best local result in each column is in \textbf{bold}.}
\label{tab:reliability_main}
\small
\setlength{\tabcolsep}{3.7pt}
\renewcommand{\arraystretch}{1.12}
\begin{tabular}{@{}lccc|ccc@{}}
\toprule
\multirow{2}{*}{\textbf{Method}} &
\multicolumn{3}{c|}{\textbf{AppRate (\%)}} &
\multicolumn{3}{c}{\textbf{UniqueRS}} \\
\cmidrule(lr){2-4}\cmidrule(l){5-7}
& @$1$ & @$5$ & @$10$ & @$1$ & @$5$ & @$10$ \\
\midrule
Stage 1               & 44.1 & 36.9 & 30.5 & 0.48 & 1.82 & 2.91 \\
Stage 2 Frozen-TE     & \textbf{63.2} & \textbf{46.5} & 35.1 & \textbf{0.69} & \textbf{2.28} & 3.33 \\
\rowcolor{softgray}
Stage 2 EMA Mid       & 59.7 & 43.9 & \textbf{34.1} & 0.65 & 2.15 & 3.21 \\
\bottomrule
\end{tabular}
\vspace{-5mm}
\end{table}

Stage 2 substantially improves applicability over Stage 1, showing that the ranking objective moves chemically executable templates closer to the head of the retrieval list. At the same time, the relationship between template-level diagnostics and final reactant accuracy is not one to one. Frozen-TE yields the highest AppRate@1, and also slightly higher template top-1 retrieval, yet EMA Mid achieves higher final reaction accuracy. The reason is that Reaction@$k$ is computed after template application, reactant canonicalization, and deduplication over the retrieved template window, not directly from the top retrieved labeled template. Templates that fail to apply under RDKit generate no reactant candidate and are therefore absent from the final deduplicated reactant ranking, so a lower ranked but applicable template can become the top ranked final reactant set. In other words, a model can retrieve a template that is not the recorded ground truth template, or retrieve non applicable templates above an applicable correct template, and still produce the correct top ranked reactant set after application, filtering, and deduplication. The yield-based diagnostics in Appendix~\ref{app:reliability_extended} support this interpretation: EMA Mid places ground-truth-producing templates more effectively within the retrieved prefix at moderate $k$, even when the labeled positive template is not retrieved first. We therefore interpret EMA's effect as improving the conversion from retrieved template sets to final reactant predictions, rather than simply maximizing labeled-template retrieval or raw applicability at rank 1; Appendix~\ref{app:metric_interpretation} discusses this metric decoupling in more detail.

We also quantify \emph{secondary} successes, where the top-1 correct reactant set is produced by a template other than the recorded positive template. Of the 3,129 correct top-1 EMA Mid predictions, 1,790 are \emph{primary} successes and 1,339 are \emph{secondary} successes, so 42.8\% of correct top-1 reactant predictions arise from a non-labeled template. This helps explain the TRetr/Reaction gap: the model often retrieves an alternative explicit template that still generates the ground-truth reactants after application and deduplication. Additional quantitative and qualitative analysis is provided in Appendix~\ref{app:secondary_success_analysis}.

\subsection{Transfer Regime: Matched-Budget Scaling Result}
\label{sec:exp_transfer}

We report the transfer setting separately because it combines USPTO-Full pretraining with USPTO-50k fine-tuning. Table~\ref{tab:transfer_main} reports the matched-budget transfer result, where fine-tuning uses the same Stage-2 candidate budget as the local models. Transfer initialization reaches 75.4\% top-1 reaction accuracy, suggesting that broader chemical coverage from large-scale pretraining can improve performance under this setup. Because the leakage audit is defined over the strict template-valid reaction space, and because this row uses Frozen-TE fine-tuning to isolate large-scale initialization, we treat it as a scaling result rather than the primary protocol-controlled comparison.

\begin{table}[t]
\centering
\caption{Matched-budget transfer result on USPTO-50k. The transfer model is initialized from USPTO-Full Stage 2 and fine-tuned with the same Stage-2 candidate budget used in the local setting.}
\label{tab:transfer_main}
\small
\setlength{\tabcolsep}{4.0pt}
\renewcommand{\arraystretch}{1.12}
\begin{tabular}{@{}lcccccc@{}}
\toprule
\multirow{2}{*}{\textbf{Method}} &
\multicolumn{4}{c}{\textbf{Reaction Acc. (\%)}} &
\multicolumn{2}{c}{\textbf{TRetr. (\%)}} \\
\cmidrule(lr){2-5}\cmidrule(l){6-7}
& Top-1 & Top-3 & Top-5 & Top-10 & Top-1 & Top-10 \\
\midrule
Transfer Frozen-TE FT ($|\mathcal{C}|{=}64$) & 75.4 & 90.8 & 93.8 & 95.9 & 74.2 & 95.3 \\
\bottomrule
\end{tabular}
\vspace{-5mm}
\end{table}

The transfer model also shows a smaller gap between template retrieval and final reactant accuracy than the local models, suggesting that large-scale pretraining better aligns top-ranked templates with final correct reactant sets. Extended transfer and USPTO-Full results are reported in Appendix~\ref{app:extended_results}.
\vspace{-3mm}
\section{Limitations and Future Work}
\label{limitations}
Our study has several limitations. First, although the local USPTO-50k regime provides the cleanest validation of the proposed method, the transfer regime still reflects the effect of larger-scale pretraining in addition to the fine-tuning method, so it should not be interpreted as a pure architecture-only gain. Second, the local ablations cover the most important design choices, but not the full hyperparameter space; in particular, KLD regularization and alternating freeze schedules may behave differently under other settings. Third, our reliability analysis shows that ranking quality and applicability are related but not identical, suggesting that explicit modeling of template applicability could further improve the usefulness of retrieved templates. Fourth, although we now quantify the prevalence of secondary successes, we do not yet provide a finer taxonomy of the chemical relation between alternative successful templates and the recorded label. Finally, while the EMA mechanism is effective on the template libraries studied here, its behavior on substantially larger libraries and under broader distribution shift remains to be tested.

Several directions follow naturally from these observations. The most immediate is to incorporate chemistry-aware supervision, for example through auxiliary losses that directly reward applicable templates and suppress high-confidence non-applicable ones. A second direction is to analyze template equivalence classes more directly, including cases where the correct reactant set is produced by a template other than the recorded positive label. A third direction is to study data-efficient transfer and cross-library adaptation more systematically, especially under matched training and evaluation conditions. A fourth direction is to extend the method beyond single-step prediction and test whether the gains in template quality, applicability, and long-tail behavior translate into stronger multi-step planning performance.
\vspace{-3mm}
\section{Conclusion}
\label{conclusion}

We presented \textbf{ConRetroBert}, a two-stage dual-encoder framework for template-based single-step retrosynthesis. The method replaces full-library template classification with contrastive product-template retrieval followed by listwise ranking over hard candidate templates, and introduces an EMA-stabilized template encoder to allow controlled template-side adaptation without destabilizing the retrieval bank.

Across local USPTO-50k experiments, Stage 2 ranking substantially improves over Stage 1 retrieval, and EMA stabilization yields a further gain over the frozen-template baseline while remaining stable across repeated runs. The long-tail and reliability analyses further support the main thesis of the paper: retrieval-based template prediction is especially well suited to the rare-template regime and produces chemically useful template candidates, not only stronger average accuracy. As a separate scaling result, larger-scale pretraining yields a substantial additional improvement after fine-tuning.

The main message is that template-based retrosynthesis need not trade away competitive performance for interpretability. With the right retrieval objective, ranking formulation, and stabilization mechanism, explicit reaction templates can remain both strong predictive objects and useful chemical actions.
\vspace{-3mm}


{
\small
\bibliographystyle{plainnat}
\bibliography{bibliography}
\medskip
}

\newpage
\appendix

\section{Technical Appendices and Supplementary Material}
\label{appendix}

This appendix provides supplementary details for reproducibility and completeness. We include related work, additional architecture and optimization details, data curation procedures, evaluation protocol specifications, and extended experimental settings that support the main paper.

\begin{figure}[!htbp]
    \centering
    \includegraphics[width=\linewidth]{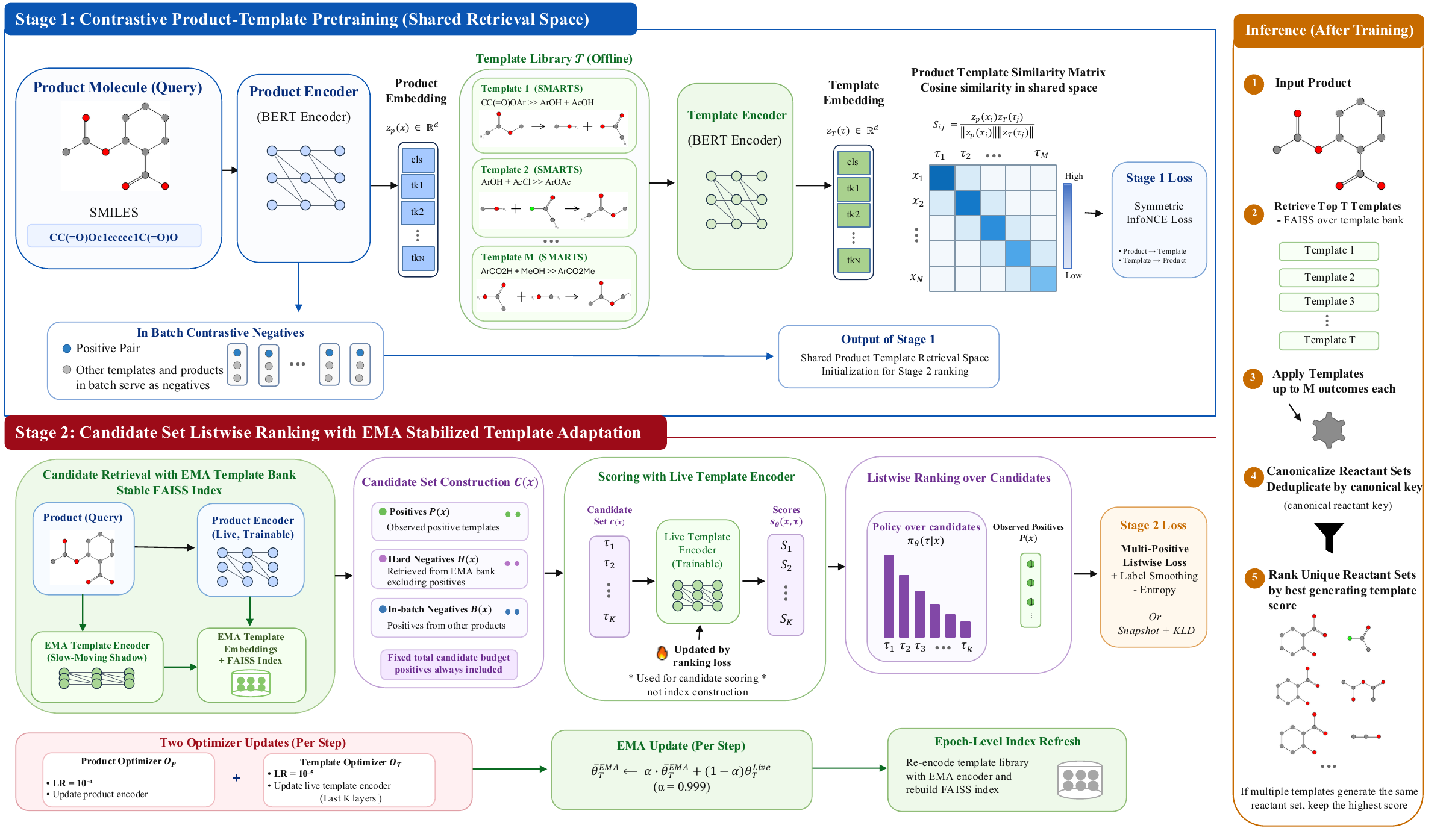}
    \caption{
    Overview of \textbf{ConRetroBert}. \textbf{Stage 1} learns a shared product--template retrieval space by contrastive pretraining of a dual encoder over product SMILES and template SMARTS. \textbf{Stage 2} refines this space into a template proposal policy through candidate-set listwise ranking over observed positives, hard negatives, and in-batch negatives. To enable template-side adaptation without destabilizing hard-negative mining, the live template encoder is trained with a separate optimizer while a slow-moving EMA template encoder is used to construct the retrieval bank and refresh the FAISS index at the epoch level. At \textbf{inference time}, the trained model retrieves top-ranked templates, applies them to the product, canonicalizes and deduplicates the resulting reactant sets, and ranks unique reactant candidates by the best generating template score.
    }
    \label{fig:conretro_pipeline}
\end{figure}

\section{Additional Method Details}
\label{app:method_details}

\subsection{Architecture and Tokenization}
\label{app:architecture}

Both product and template encoders use the same Transformer architecture but do not share parameters. Each tower has hidden dimension $d = 256$, $L = 6$ Transformer layers, $H = 8$ attention heads, dropout $0.1$, and maximum sequence length $L_{\max} = 384$. The tokenizer is a shared character level vocabulary built from product SMILES and template SMARTS strings. The sequence representation is taken directly from the first token hidden state of each encoder, without an additional projection head. These vectors are then $\ell_2$ normalized before scoring.

\subsection{Candidate Set Construction}
\label{app:candidate_construction}

For each product $x$, Stage~2 constructs a fixed-size candidate list for the listwise denominator. We write the candidate source as
\begin{equation}
\mathcal{C}_{\mathrm{raw}}(x)
=
\bigl[P(x),\,\mathcal{B}(x),\,\mathcal{H}(x),\,\mathcal{R}(x)\bigr],
\end{equation}
where $P(x)$ is the set of observed positive templates, $\mathcal{B}(x)$ contains in-batch negatives from positives associated with other products in the same minibatch, $\mathcal{H}(x)$ contains FAISS-mined hard negatives, and $\mathcal{R}(x)$ is used only as a random-fill fallback.

Positives are always inserted first. We then add up to 8 in-batch negatives after skipping templates that overlap with $P(x)$ or have already been selected. Hard negatives are drawn from the FAISS top-$128$ retrieved templates: we scan this ranked list in order and add eligible templates until the hard-negative quota is filled, again skipping templates that overlap with $P(x)$ or have already been selected. Thus, because $|P(x)|$ can be larger than one, the exact number of hard negatives included before the final size correction may vary by product.

In all Stage~2 experiments, the final listwise denominator has fixed size $|\mathcal{C}(x)|=64$. If the ordered candidate list exceeds 64 entries, we keep the first 64 entries and discard candidates from the end, so observed positives are prioritized, followed by in-batch negatives and then FAISS hard negatives. If the list remains below 64 after scanning the eligible FAISS top-$128$ pool and adding in-batch negatives, the remaining slots are filled with randomly sampled templates. If this random-fill step still cannot produce enough eligible distinct templates, existing candidates are sampled with replacement as a final fallback to preserve a fixed tensor shape and softmax denominator size. This replacement fallback affects only padding and does not introduce additional distinct templates.

The frozen-template-encoder baseline uses this procedure with 8 in-batch negatives and no random negatives unless the FAISS top-$128$ eligible pool is insufficient. The EMA and Snapshot+KLD variants use the same final candidate-set size, with adjustments for memory-safe live template scoring described in Appendix~\ref{app:training_config}.

The unified Stage~2 algorithm and the mathematical rationale for these design choices are provided in Appendix~\ref{app:stage2_algorithm} and Appendix~\ref{app:method_math}.
\subsection{Label Smoothing}
\label{app:label_smoothing}

The main text writes the Stage 2 loss as a multi positive listwise objective. In implementation, we use label smoothing with $\epsilon=0.02$. Let $C=|\mathcal{C}(x)|$ and $m=|P(x)|$. The smoothed target distribution is
\begin{equation}
\tilde{y}_{\tau}
=
\begin{cases}
\frac{1-\epsilon}{m}+\frac{\epsilon}{C},
& \tau\in P(x),\\[4pt]
\frac{\epsilon}{C},
& \tau\notin P(x).
\end{cases}
\end{equation}
The smoothed ranking loss is
\begin{equation}
\mathcal{L}_{\mathrm{smooth}}(x)
=
-\sum_{\tau\in \mathcal{C}(x)}
\tilde{y}_{\tau}
\log \pi_\theta(\tau\mid x).
\end{equation}

The final Stage 2 objective used in the reported ranking experiments combines this
label-smoothed loss with the entropy regularizer from Eq.~\ref{eq:stage2_full}:
\begin{equation}
\mathcal{L}_{\mathrm{S2}}(x)
=
\mathcal{L}_{\mathrm{smooth}}(x)
-
\beta H(\pi_{\theta}(\cdot \mid x)).
\end{equation}
We use \(\epsilon = 0.02\) and \(\beta = 0.001\) in the reported Stage 2 experiments.

\subsection{Optimizer Separation}
\label{app:optimizer_separation}

The EMA Stage 2 model uses two AdamW optimizers. The product optimizer $\mathcal{O}_P$ updates the product encoder with learning rate $10^{-4}$ and a warmup cosine schedule. The template optimizer $\mathcal{O}_T$ updates only the selected trainable part of the template encoder with learning rate $10^{-5}$. In the main local EMA configuration, only the last three template encoder layers and the final normalization parameters are trainable. Lower layers, token embeddings, and positional embeddings are frozen.

This design follows the different roles of the two towers. The product encoder acts as the query network and should adapt quickly to improve ranking. The template encoder defines the retrieval basis used to construct hard negatives and must move slowly enough that the index remains meaningful throughout an epoch. Separate optimizers also isolate AdamW moment estimates, which is useful when the template encoder is frozen or unfrozen in alternating variants.

\subsection{EMA Index Refresh}
\label{app:ema_refresh}

Let $\theta_T^{(t)}$ be the live template encoder parameters after step $t$ and let $\bar{\theta}_T^{(t)}$ denote the EMA shadow. After each live template update, the shadow is updated as
\begin{equation}
\bar{\theta}_T^{(t)}
=
\alpha \bar{\theta}_T^{(t-1)}
+
(1-\alpha)\theta_T^{(t)}.
\end{equation}
With $\alpha=0.999$, the EMA half life is approximately $693$ gradient steps. At the start of each epoch, all templates are re encoded with $f_T(\cdot;\bar{\theta}_T)$ and the FAISS index is rebuilt. Hard negatives for that epoch are then mined from the refreshed index.

\subsection{KLD Regularization}
\label{app:kld_regularization}

Some ablations replace EMA stabilization with a KLD-based teacher regularization scheme. In these runs, the live template encoder is regularized toward a snapshot teacher distribution rather than an EMA teacher distribution. Let \(\hat{\theta}_T\) denote the teacher parameters frozen at the start of the current epoch. The teacher and live candidate-set policies are denoted by \(\pi^{\mathrm{snap}}_{\theta}(\cdot\mid x)\) and \(\pi^{\mathrm{live}}_{\theta}(\cdot\mid x)\), respectively. We use the forward distillation direction from the snapshot teacher to the live model:
\begin{equation}
\mathcal{L}_{\mathrm{KLD}}
=
\mathrm{KL}
\left(
\pi^{\mathrm{snap}}_\theta(\cdot\mid x)
\;\Vert\;
\pi^{\mathrm{live}}_\theta(\cdot\mid x)
\right).
\end{equation}
The full objective becomes
\begin{equation}
\mathcal{L}(x)
=
\mathcal{L}_{\mathrm{S2}}(x)
+
\lambda_{\mathrm{KLD}}\mathcal{L}_{\mathrm{KLD}}(x).
\end{equation}
We use \(\lambda_{\mathrm{KLD}}=0.1\) in the reported KLD ablations. In local USPTO-50k experiments, this variant is competitive but does not outperform EMA Mid.

\subsection{Alternating Freeze Variant}
\label{app:alternating_freeze}

The alternating variant freezes and unfreezes the two towers in cycles. During product update epochs, the template encoder is frozen and only the product encoder is trained. During template update epochs, the product encoder is frozen and the selected template encoder layers are updated. In the reported transfer variant, the template encoder is frozen for $N_f=10$ epochs and unfrozen for $N_u=2$ epochs in each cycle. The template optimizer state is reset when entering a template update phase.

\subsection{Unified Stage 2 Algorithm}
\label{app:stage2_algorithm}

Algorithm~\ref{alg:stage2} summarizes the Stage 2 training procedure in a unified form. Different reported variants are recovered by choosing different settings of the five binary flags $(\delta_e,\delta_s,\delta_k,\delta_a,\delta_f)$, corresponding respectively to EMA teacher, snapshot teacher, KLD regularization, alternating updates, and frozen template encoder. For clarity, the reported KLD based ablations use snapshot teacher regularization rather than EMA, so $\delta_k$ is activated only together with $\delta_s$, not with $\delta_e$.

\begin{algorithm}[h]
\caption{Unified Stage 2 Dual Encoder Training}
\label{alg:stage2}
\small
\begin{algorithmic}[1]
\Require Stage 1 initialized encoders $\theta_P^{(0)}, \theta_T^{(0)}$, template library $\mathcal{T}$, epochs $E$
\Require Flags $\delta_e$ (EMA teacher), $\delta_s$ (snapshot teacher), $\delta_k$ (KLD regularization), $\delta_a$ (alternating updates), $\delta_f$ (frozen template encoder)
\Require Hyperparameters $\tau_{\mathrm{temp}}, \lambda_{\mathrm{KLD}}, \alpha, N_f, N_u$
\State Initialize product optimizer $\mathcal{O}_P$ and template optimizer $\mathcal{O}_T$
\If{$\delta_e = 1$}
    \State $\bar{\theta}_T \gets \theta_T^{(0)}$
\EndIf
\If{$\delta_s = 1$}
    \State $\hat{\theta}_T \gets \theta_T^{(0)}$
\EndIf
\State Build retrieval index $\mathcal{I}$ from $\bar{\theta}_T$ if $\delta_e=1$, else $\hat{\theta}_T$ if $\delta_s=1$, else $\theta_T$
\For{$e = 1,\dots,E$}
    \If{$\delta_a = 1$}
        \State Determine whether product or template tower is frozen using $(N_f,N_u)$
    \Else
        \State Freeze template tower if and only if $\delta_f = 1$
    \EndIf
    \If{$\delta_s = 1$}
        \State $\hat{\theta}_T \gets \theta_T$
    \EndIf
    \State Rebuild retrieval index from teacher parameters for the current epoch
    \For{each minibatch $\mathcal{B}$}
        \For{each product $x_i$ in $\mathcal{B}$}
            \State Construct candidate set $\mathcal{C}(x_i)=P(x_i)\cup\mathcal{H}(x_i)\cup\mathcal{B}(x_i)$ 
            \State Compute scores $s_\theta(x_i,\tau)$ for $\tau \in \mathcal{C}(x_i)$
            \State Compute listwise loss $\mathcal{L}_{\mathrm{rank}}(x_i)$
            \If{$\delta_k = 1$ and a teacher distribution is available}
                \State Add $\lambda_{\mathrm{KLD}}\mathcal{L}_{\mathrm{KLD}}(x_i)$
            \EndIf
        \EndFor
        \State Update $\theta_P$ if product encoder is trainable
        \State Update $\theta_T$ if template encoder is trainable
        \If{$\delta_e = 1$ and template encoder is trainable}
            \State $\bar{\theta}_T \gets \alpha \bar{\theta}_T + (1-\alpha)\theta_T$
        \EndIf
    \EndFor
\EndFor
\end{algorithmic}
\end{algorithm}

\subsection{Stability of the EMA Retrieval Bank}
\label{app:method_math}
 
The role of the EMA template encoder in ConRetroBert is to smooth the evolution
of the retrieval bank used for candidate construction across epochs.
We formalize a limited but honest notion of stability that is directly tied to
the training mechanism used in the paper.  We do \emph{not} claim convergence
of the full two-encoder dynamics; the result below concerns only the
epoch-to-epoch drift of the \emph{EMA shadow bank}, not the live encoder.
 
\paragraph{Setup and notation.}
Let $\theta_T^{(t)}$ denote the live template encoder parameters after
gradient step $t$, and let $\bar\theta_T^{(t)}$ denote the EMA shadow after
the same step, updated without gradients via
\begin{equation}
  \bar\theta_T^{(t+1)}
  \;=\;
  \alpha\,\bar\theta_T^{(t)}
  \;+\;
  (1-\alpha)\,\theta_T^{(t+1)},
  \qquad \alpha = 0.999.
  \label{eq:ema_update2}
\end{equation}
The retrieval bank at the start of epoch $e$ is built from the EMA snapshot
$\bar\theta_T^{(e)}$ and is held \emph{fixed} for the entire epoch:
\begin{equation}
  \mathcal{I}^{(e)} \;=\; \mathrm{BuildIndex}\!\bigl(\bar\theta_T^{(e)},\,\mathcal{T}\bigr).
\end{equation}
For analysis, and for a fixed product query representation of product $x$, define the full-library retrieval distribution induced by bank $e$ as
\begin{equation}
  q^{(e)}(\tau\mid x)
  \;=\;
  \frac{\exp\!\bigl(s(x,\tau;\,\bar\theta_T^{(e)})\bigr)}
       {\sum_{\tau'\in\mathcal{T}}\exp\!\bigl(s(x,\tau';\,\bar\theta_T^{(e)})\bigr)}.
\end{equation}
We use the term \emph{epoch-stable} to refer to the controlled drift of
$\sup_x \bigl\|q^{(e+1)}(\cdot\mid x)-q^{(e)}(\cdot\mid x)\bigr\|_1$
under the bounded update assumptions below.
 
\paragraph{Assumptions.}
\begin{enumerate}
  \item[(i)] \textbf{Bounded live updates.}
        Each template-side gradient step satisfies
        
        $\|\theta_T^{(t+1)}-\theta_T^{(t)}\|\le \eta_T G_T$
        
        for a per-step gradient norm bound $G_T$ and template learning rate
        $\eta_T$.
  \item[(ii)] \textbf{Lipschitz score function.}
        For every product $x$ and template $\tau$,
        
        $|s(x,\tau;\theta)-s(x,\tau;\theta')|\le L_s\|\theta-\theta'\|$.
  \item[(iii)] \textbf{Lipschitz softmax.}
        There exists $L_{\mathrm{sm}}$ such that
        
        $\|\mathrm{softmax}(u)-\mathrm{softmax}(v)\|_1\le L_{\mathrm{sm}}\|u-v\|_\infty$.
\end{enumerate}
 
\paragraph{Proposition (EMA bank drift bound).}
\textit{Suppose an epoch contains $m$ template-side gradient steps. For this epoch,
reset the within-epoch step index so that the initial live parameters are
$\theta_T^{(0)}$ and the initial EMA shadow is
$\bar\theta_T^{(0)}=\bar\theta_T^{(e)}$.
Then the EMA shadow drift over one epoch satisfies}
\begin{equation}
  \bigl\|\bar\theta_T^{(m)} - \bar\theta_T^{(0)}\bigr\|
  \;\le\;
  (1-\alpha)\sum_{j=1}^{m}\alpha^{m-j}
  \bigl\|\theta_T^{(j)}-\bar\theta_T^{(0)}\bigr\|.
  \label{eq:ema_drift_exact}
\end{equation}
\textit{Using the bounded-update assumption, the live encoder satisfies
$\|\theta_T^{(j)}-\theta_T^{(0)}\|\le j\,\eta_T G_T$.
Therefore, the right-hand side of \eqref{eq:ema_drift_exact} is bounded by}
\begin{equation}
  \bigl\|\bar\theta_T^{(m)} - \bar\theta_T^{(0)}\bigr\|
  \;\le\;
  (1-\alpha)\sum_{j=1}^{m}\alpha^{m-j}
  \!\left(j\,\eta_T G_T + \|\theta_T^{(0)}-\bar\theta_T^{(0)}\|\right).
  \label{eq:ema_drift_bound}
\end{equation}
\textit{Consequently, the retrieval distribution drift satisfies}
\begin{equation}
  \sup_x
  \bigl\|q^{(e+1)}(\cdot\mid x) - q^{(e)}(\cdot\mid x)\bigr\|_1
  \;\le\;
  L_{\mathrm{sm}} L_s
  \cdot
  (1-\alpha)\sum_{j=1}^{m}\alpha^{m-j}
  \!\left(j\,\eta_T G_T + \Delta_0\right),
  \label{eq:bank_drift}
\end{equation}
\textit{where $\Delta_0 = \|\theta_T^{(0)}-\bar\theta_T^{(0)}\|$ is the
live-shadow gap at the start of the epoch.}
 
\paragraph{Proof.}
 
\textbf{Step 1: Unrolling the EMA recursion.}
Applying \eqref{eq:ema_update2} repeatedly from step $0$ to step $m$:
\begin{align}
  \bar\theta_T^{(m)}
  &= \alpha^m\,\bar\theta_T^{(0)}
     + (1-\alpha)\sum_{j=1}^{m}\alpha^{m-j}\,\theta_T^{(j)}.
  \label{eq:ema_unroll}
\end{align}
Subtracting $\bar\theta_T^{(0)}$ and using
$\alpha^m + (1-\alpha)\sum_{j=1}^m\alpha^{m-j}=1$:
\begin{align}
  \bar\theta_T^{(m)} - \bar\theta_T^{(0)}
  &= (1-\alpha)\sum_{j=1}^{m}\alpha^{m-j}
     \!\bigl(\theta_T^{(j)}-\bar\theta_T^{(0)}\bigr).
  \label{eq:shadow_diff}
\end{align}
Taking norms and applying the triangle inequality yields
\eqref{eq:ema_drift_exact}.
 
\textbf{Step 2: Bounding $\|\theta_T^{(j)}-\bar\theta_T^{(0)}\|$.}
By the triangle inequality,
\begin{equation}
  \|\theta_T^{(j)}-\bar\theta_T^{(0)}\|
  \le
  \|\theta_T^{(j)}-\theta_T^{(0)}\| + \|\theta_T^{(0)}-\bar\theta_T^{(0)}\|
  \le j\,\eta_T G_T + \Delta_0,
\end{equation}
where the second inequality uses assumption (i) applied $j$ times and the
definition $\Delta_0=\|\theta_T^{(0)}-\bar\theta_T^{(0)}\|$.
Substituting into \eqref{eq:ema_drift_exact} gives \eqref{eq:ema_drift_bound}.
 
\textbf{Step 3: Propagating to retrieval distributions.}
By assumption (ii),
\begin{equation}
  \sup_{x,\tau}
  \bigl|s(x,\tau;\bar\theta_T^{(m)})-s(x,\tau;\bar\theta_T^{(0)})\bigr|
  \;\le\;
  L_s\,\bigl\|\bar\theta_T^{(m)}-\bar\theta_T^{(0)}\bigr\|.
\end{equation}
By assumption (iii), the $\ell_1$ change in the retrieval distribution is
bounded by $L_{\mathrm{sm}}$ times the $\ell_\infty$ change in the score
vector, giving \eqref{eq:bank_drift}. $\square$
 
\paragraph{Remark: interpretation of the EMA drift bound.}
The analogous worst-case drift bound for the live encoder over \(m\) steps is
\(m\,\eta_T G_T\).  Ignoring the initial live-shadow gap \(\Delta_0\), the
dominant EMA drift term in \eqref{eq:ema_drift_bound} is
\begin{equation}
  (1-\alpha)\sum_{j=1}^{m}\alpha^{m-j}\,j\,\eta_T G_T
  =
  \frac{m-(m+1)\alpha+\alpha^{m+1}}{1-\alpha}\,\eta_T G_T .
  \label{eq:ema_vs_live_closed}
\end{equation}
This expression is at most \(m\,\eta_T G_T\), but it remains \(O(m)\) in the
deterministic worst case.  Thus EMA should not be interpreted as preventing
linear epoch-scale drift when live template updates move coherently in one
direction.  For example, in a steady state where the live encoder drifts at a
constant vector \(g\) per step, the EMA shadow lags the live encoder by
approximately \(\alpha(1-\alpha)^{-1}\|g\|\), and the EMA shadow itself moves
at the same asymptotic rate \(\|g\|\) per step.
 
The practical stabilizing effect of EMA is therefore not a change in the
deterministic worst-case order in \(m\), but a smoothing of high-frequency
variation in the template encoder before the full template library is re-encoded
and the FAISS index is rebuilt.  The index used for hard-negative mining is constructed from
a slow-moving average of recent live encoders rather than from the instantaneous
live encoder.  This makes candidate construction less sensitive to transient
optimizer noise and abrupt step-to-step fluctuations, while still allowing the
template representation to adapt through the live encoder.
 
\paragraph{Scope.}
This result is intentionally limited.
It does not prove convergence of the full two-encoder training dynamics, nor
does it characterize the exact change in the discrete top-$k$ candidate set
(which depends on score \emph{differences}, not absolute score magnitudes).
What it establishes is that, holding the product query representation fixed and
under bounded live template updates, the distribution induced by the EMA
retrieval bank changes in a controlled way governed by the EMA recursion and
the live-shadow gap; empirically, this provides a smoother basis for candidate
construction than using the instantaneous live encoder.

\subsection{Model Configuration Summary}
\label{app:model_config_summary}

\begin{table}[H]
\centering\small
\caption{Core architectural configuration of ConRetroBert. Parameter counts are reported for the Stage-2 EMA Mid and Stage-2 Frozen-TE variants. In EMA Mid, the product encoder is fully trainable, while only the last three transformer layers and final normalization of the template encoder are trainable.}
\label{tab:model_config_summary}
\renewcommand{\arraystretch}{1.12}
\setlength{\tabcolsep}{5pt}
\begin{tabular}{@{}lcc@{}}
\toprule
\textbf{Item} & \textbf{Stage-2 EMA Mid} & \textbf{Stage-2 Frozen-TE} \\
\midrule
Architecture & Dual BERT style encoder & Dual BERT style encoder \\
Product encoder & Trainable & Trainable \\
Template encoder & Last 3 layers + final norm trainable & Frozen \\
Tokenizer & Shared character tokenizer & Shared character tokenizer \\
Tokenizer vocab size & 75 & 75 \\
Maximum sequence length & 384 & 384 \\
Encoder layers & 6 per encoder & 6 per encoder \\
Hidden size & 256 & 256 \\
Attention heads & 8 & 8 \\
Feed forward size & 1024 & 1024 \\
Dropout & 0.1 & 0.1 \\
Positional embeddings & Learned & Learned \\
Projection head & None & None \\
Sequence representation & First token hidden state & First token hidden state \\
Similarity & Dot product of normalized embeddings & Dot product of normalized embeddings \\
Parameters per encoder & 4,856,576 & 4,856,576 \\
Total parameters & 9,713,152 & 9,713,152 \\
Active trainable parameters & 7,226,368 & 4,856,576 \\
Non trainable parameters & 2,486,784 & 4,856,576 \\
\bottomrule
\end{tabular}
\end{table}

\subsection{Training Configuration}
\label{app:training_config}

\paragraph{Stage 1.}
Contrastive pretraining uses batch size $256$, AdamW ($\mathrm{lr}=10^{-4}$, weight decay $10^{-2}$), mixed precision training, temperature $\tau=0.07$, and a warmup cosine schedule with $500$ warmup steps over $100$ epochs.

\paragraph{Stage 2 with Frozen Template Encoder.}
Frozen template encoder training uses batch size $256$, AdamW ($\mathrm{lr}=2\times10^{-4}$, weight decay $10^{-2}$), temperature $\tau=0.07$, entropy term $\beta=0.001$, and label smoothing $\epsilon=0.02$. Template embeddings are precomputed once from the frozen template encoder and held fixed throughout training.

\paragraph{Stage 2 Variants with Trainable Template Encoders.}
All Stage 2 variants with trainable template encoders use batch size $64$ with two AdamW optimizers: a product optimizer ($\mathrm{lr}=10^{-4}$, warmup cosine) and a template optimizer ($\mathrm{lr}=10^{-5}$, fixed). At each step, the live product encoder encodes all $64$ products and assembles per product candidate sets from the retrieval bank. Candidates are then scored with the live template encoder: all $B\times C$ candidate IDs are globally deduplicated, the unique set $\mathcal{U}$ is forwarded through the template encoder, and embeddings are gathered back through inverse indices so that template encoder gradients are computed without redundant forward passes. The $64$ products are processed in micro batches of $16$, while the deduplicated template embeddings are reused across all chunks. Gradients are accumulated over $4$ consecutive data batches before each optimizer step, giving an effective batch size of $256$. The loss is scaled by $1/4$, and gradient clipping is applied with $\|\nabla\|_2 \leq 1.0$.

\section{Evaluation Protocol}
\label{app:evaluation_protocol}

\subsection{From Template Rankings to Reactant Rankings}
\label{app:template_to_reactant}

The model ranks templates, but standard single-step retrosynthesis accuracy is computed over reactant sets. We therefore evaluate through template application. For each product \(x\), we first retrieve a large FAISS pool from the precomputed template bank, re-rank that pool with the model score to form the evaluation candidate set, and then apply only the top-ranked templates within that re-ranked set. For each applied template, we collect up to \(M\) outcomes and canonicalize each predicted reactant set. A reactant set is converted into a canonical key by parsing all molecules with RDKit, converting them to canonical isomeric SMILES, sorting molecular components lexicographically, and joining them with a period.

\subsection{Deduplication}
\label{app:deduplication}

Different templates or different match sites can generate the same reactant set. For each unique canonical reactant key $r$, we retain the best template score:
\begin{equation}
\mathrm{score}(r)
=
\max_{\tau:\, r\in \mathrm{Apply}(x,\tau)}
s_\theta(x,\tau).
\end{equation}
Unique reactant keys are then sorted by this score. Exact-match top-$k$ accuracy is computed by checking whether the canonical ground-truth reactant key appears among the top-$k$ predicted keys.

\subsection{Metrics}
\label{app:metrics}

We report two primary benchmark metrics and four diagnostic metrics.

\paragraph{Reaction top-$k$ accuracy.}
The fraction of test products for which the canonical ground-truth reactant set appears among the top-$k$ deduplicated reactant predictions.

\paragraph{Template retrieval top-$k$ accuracy.}
The fraction of test products for which an observed positive template appears among the top-$k$ retrieved templates before template application.

\paragraph{Applicability rate (AppRate@$k$).}
The fraction of retrieved top-$k$ template slots that successfully apply to the product and produce at least one reaction outcome under RDKit template application.

\paragraph{Unique reactant sets (UniqueRS@$k$).}
The mean number of distinct canonicalized reactant sets generated by the top-$k$ retrieved templates.

\paragraph{Yield coverage (YieldCov@$k$).}
The fraction of products for which at least one of the top-$k$ retrieved templates generates the ground-truth reactant set.

\paragraph{Yield rate (YieldRate@$k$).}
The fraction of top-$k$ retrieved template slots, aggregated across products, that generate the ground-truth reactant set.

This denominator policy is used consistently for all main-paper and appendix result tables.

\subsection{Interpreting Template Level Diagnostics vs.\ Final Reactant Accuracy}
\label{app:metric_interpretation}

The template level diagnostics and the final reaction accuracy measure different parts of the pipeline and should not be interpreted as interchangeable quantities. In particular, Template retrieval top-$k$ asks whether the recorded positive template appears in the retrieved prefix, while Reaction top-$k$ is computed only after template application, canonicalization, reactant set deduplication, and final reactant ranking. A model can therefore miss the labeled positive template at rank 1, or retrieve a rank 1 template with lower raw applicability, and still produce the correct top ranked reactant set after application and deduplication.

The same distinction applies to YieldCov@$k$ and YieldRate@$k$. YieldCov@$k$ measures whether at least one of the top-$k$ retrieved template slots produces the ground truth reactant set, whereas Reaction Acc@$k$ is computed from the final deduplicated reactant-set ranking built from the larger evaluation window. Therefore YieldCov@$k$ is not an upper bound on Reaction Acc@$k$ at small $k$.

This distinction helps explain the apparent decoupling observed in the local regime. EMA Mid is slightly below Frozen TE on Template retrieval@1 and AppRate@1, yet it achieves higher final Reaction Acc@1. The yield diagnostics indicate why: EMA Mid places ground truth producing templates more effectively within the retrieved prefix at moderate $k$, with YieldCov@5 of $67.13\%$ versus $64.76\%$ for Frozen TE, YieldCov@10 of $75.07\%$ versus $73.15\%$, and YieldCov@20 of $82.31\%$ versus $80.40\%$. Similarly, EMA Mid has slightly lower YieldRate@1 than Frozen TE, $42.08\%$ versus $44.12\%$, but slightly higher YieldRate at $k=3,5,10,$ and $20$. This pattern is consistent with a model whose retrieved template set supports better final reactant ranking after application and deduplication, even when the recorded labeled template is not retrieved first.

\subsection{Filtered vs.\ Unfiltered Evaluation}
\label{app:filtered_eval}

To clarify the effect of template-validity filtering at test time, we evaluated the same Stage-2 EMA checkpoint on two USPTO-50k test conditions: the standard unfiltered benchmark split and a valid-only subset retaining only records whose recorded ground-truth template can be reapplied to exactly regenerate the recorded ground-truth reactants. The valid-only subset is therefore easier by construction, because it excludes examples for which the recorded template-action pair is internally inconsistent under the reaction engine.

Across all reported cutoffs, the valid-only condition is slightly higher than the unfiltered benchmark, with gains of roughly 0.9--1.2 points in exact precursor accuracy. We therefore use the unfiltered USPTO-50k result for comparison with prior reported benchmarks, and interpret the valid-only numbers only as a diagnostic of evaluation noise introduced by invalid ground-truth template records.

\begin{table}[H]
\centering\small
\caption{Stage-2 EMA evaluation on standard unfiltered USPTO-50k and a valid-only filtered subset. The best EMA Mid checkpoint is used in both cases. $\Delta$ denotes filtered minus unfiltered.}
\label{tab:filtered_vs_unfiltered}
\renewcommand{\arraystretch}{1.12}
\setlength{\tabcolsep}{4.0pt}
\begin{tabular}{@{}lcccccc@{}}
\toprule
\textbf{Condition} & \textbf{@1} & \textbf{@3} & \textbf{@5} & \textbf{@10} & \textbf{@20} \\
\midrule
Unfiltered   & 62.52 & 81.64 & 85.43 & 87.89 & 88.73 \\
Valid-only    & 63.42 & 82.76 & 86.58 & 89.06 & 89.88 \\
$\Delta$      & +0.90 & +1.12 & +1.15 & +1.17 & +1.15 \\
\bottomrule
\end{tabular}
\end{table}

\section{Data Curation}
\label{app:data_curation}

\subsection{Datasets}
\label{app:datasets}

We use USPTO-50k as the primary benchmark and USPTO-Full as the large scale source for transfer pretraining. Both datasets are passed through the same chemistry aware curation pipeline before training or evaluation. This pipeline distinguishes three progressively stricter stages: successful template extraction, multi-input template detection, and strict forward validation by reapplying the extracted template to the product and checking whether it regenerates the recorded reactants.

Table~\ref{tab:dataset_preprocessing_stats} summarizes these preprocessing statistics. For USPTO-50k, successful template extraction is nearly lossless, and the main reduction comes from strict validation. For USPTO-Full, extraction remains high but the stricter validation stage removes a larger number of reactions, yielding the final \emph{valid reaction} subset used for transfer pretraining before leakage removal.

\begin{table}[!hbt]
\centering
\caption{Dataset preprocessing and template validation statistics.}
\label{tab:dataset_preprocessing_stats}
\resizebox{\textwidth}{!}{%
\begin{tabular}{llrrrrr}
\toprule
\textbf{Dataset} & \textbf{Split} & \textbf{Quantity} & \textbf{Successful Template Extraction} &
\textbf{Multi-Input Templates} & \textbf{Forward Validation} & \textbf{Valid Reaction} \\
\midrule
USPTO-50k  & train & 40{,}008  & 39{,}992  & 322      & 680      & 38{,}990 \\
USPTO-50k  & val   & 5{,}001   & 5{,}001   & 34       & 92       & 4{,}875 \\
USPTO-50k  & test  & 5{,}007   & 5{,}005   & 36       & 79       & 4{,}890 \\
\midrule
USPTO-Full & train & 810{,}496 & 768{,}067 & 50{,}286 & 18{,}647 & 699{,}134 \\
USPTO-Full & val   & 101{,}311 & 96{,}070  & 6{,}383  & 2{,}347  & 87{,}340 \\
USPTO-Full & test  & 101{,}311 & 95{,}924  & 6{,}315  & 2{,}328  & 87{,}281 \\
\bottomrule
\end{tabular}%
}
\end{table}

For the local USPTO-50k setting, the reported benchmark evaluation continues to use the standard unfiltered test split, while strict valid only subsets are used as diagnostic artifacts in the curation pipeline. For the transfer setting, we use the USPTO-Full \emph{valid reaction} training subset as the starting point, and then remove all reactions that overlap with USPTO-50k validation or test by canonical reaction signature matching.

\subsection{Template Validity Filtering}
\label{app:template_validity}

Template-based learning requires more than syntactic extraction of a SMARTS transformation. An extracted template must also be chemically usable under the single product evaluation protocol. We therefore distinguish between three stages of validity.

First, a reaction must admit \emph{successful template extraction}. Reactions that fail template extraction are excluded immediately because they do not provide a usable template target.

Second, we identify \emph{multi-input templates}. These are templates whose product side contains multiple disconnected components and therefore requires multiple molecular inputs during application. Under single step retrosynthesis, template application begins from one product molecule, so such templates are not directly usable in the evaluation setting.

Third, we perform \emph{strict forward validation}. The extracted template is applied back to the recorded product and is retained only if it regenerates the recorded reactants under strict canonical matching. Reactions that survive this step form the \emph{valid reaction} subset. In our pipeline, these valid reaction files are the highest confidence artifacts for template grounded training and curation.

\subsection{Multi-Input Template Handling}
\label{app:multi_input}

Some extracted SMARTS transformations contain multiple product side fragments. In RDKit terms, these correspond to templates with more than one required reactant input during application. Since single step retrosynthesis evaluation starts from a single product molecule, such templates would otherwise return empty outputs and be incorrectly conflated with ordinary retrieval failures.

We therefore handle multi-input templates explicitly at two levels. First, they are tracked during preprocessing as a separate failure mode, rather than being merged into generic not applicable outcomes. Second, template libraries used in model training and retrieval are filtered to the single-input setting required by the benchmark protocol. If a test example still yields no prediction because all retrieved templates are unsupported under this single-input setting, that row is counted as a miss in the reported benchmark denominator. This policy is consistent with the evaluation protocol described in Appendix~\ref{app:evaluation_protocol}.

\subsection{Cross Dataset Leakage Removal}
\label{app:leakage}

Since USPTO-50k is derived from the broader USPTO reaction corpus, transfer
pretraining on USPTO-Full can leak downstream validation or test chemistry
unless the pretraining corpus is explicitly decontaminated. Raw string matching
is not sufficient because the same chemistry can appear with different atom
mapping, reactant ordering, or equivalent SMILES forms. We therefore compute
canonical reaction signatures by stripping atom maps, canonicalizing product and
reactant SMILES with RDKit, and sorting reactant components lexicographically
before comparison.

Leakage removal is performed after strict validation of the USPTO-Full
pretraining corpus. This choice follows the template-grounded nature of the
model: USPTO-Full pretraining uses only reactions with extractable, single-input,
applicable templates that strictly regenerate the recorded reactants. Concretely,
we begin from the USPTO-Full train \emph{valid reaction} subset containing
699{,}134 reactions, and remove every reaction whose canonical signature matches
any reaction in the USPTO-50k validation or test valid reaction subsets.
Table~\ref{tab:leakage_stats} reports the overlap counts and the size of the
final leakage-cleaned pretraining corpus.

\begin{table}[!hbt]
\centering
\caption{Cross dataset leakage removal from USPTO-Full train using canonical reaction signatures against USPTO-50k validation and test valid reaction subsets.}
\label{tab:leakage_stats}
\small
\renewcommand{\arraystretch}{1.12}
\setlength{\tabcolsep}{5pt}
\begin{tabular}{lrrr}
\toprule
\textbf{Reference set} & \textbf{Reference rows} & \textbf{Overlap signatures in train} & \textbf{Count} \\
\midrule
USPTO-50k test & 4{,}890 & 2{,}706 & 2{,}866 removed \\
USPTO-50k val  & 4{,}875 & 2{,}692 & 2{,}863 removed \\
\midrule
Combined union removed & & & 5{,}722 \\
Leakage-cleaned USPTO-Full train & & & 693{,}412 \\
\bottomrule
\end{tabular}
\end{table}

The sum of removals against validation and test separately is slightly larger
than the final union removed count because a small number of USPTO-Full training
reactions overlap with both USPTO-50k validation and test simultaneously. After
union-based deduplication, the final leakage-cleaned transfer corpus contains
693{,}412 reactions. This is the corpus used for all reported transfer
experiments.

The valid-reaction reference was not selected to make the benchmark easier.
The standard USPTO-50k validation and test splits remain unfiltered at
evaluation time, so reactions that fail strict template validation are still
included in the denominator and can lower the reported reaction accuracy when
the model does not recover the recorded reactants. Thus, the decontamination is
exact with respect to the template-valid reaction space used for USPTO-Full
pretraining, while the evaluation remains on the standard unfiltered benchmark
splits. We nevertheless note that a stronger audit would additionally compare
against all canonicalizable rows in the complete unfiltered validation and test
splits, including rows outside the strict valid template reference. We therefore
treat USPTO-Full initialization as a scaling result rather than as the primary
protocol-controlled comparison; the primary method validation remains the local
USPTO-50k regime.

\section{Additional Experimental Details}
\label{app:exp_details}

\subsection{Training Regimes}
\label{app:exp_regimes_appendix}

\paragraph{Local USPTO-50k regime.}
All stages are trained using USPTO-50k only. This is the primary regime used to validate the proposed method.

\paragraph{Transfer regime.}
A USPTO-Full Stage-2 checkpoint is used as initialization and then fine-tuned on USPTO-50k. This regime is reported separately because it reflects both larger-scale pretraining and the proposed fine-tuning method.

\subsection{Variant Definitions}
\label{app:variant_definitions}

\paragraph{Stage 1.}
Contrastively pretrained dual encoder without Stage-2 ranking refinement.

\paragraph{Stage 2 Frozen-TE.}
Stage-2 listwise ranking with a fixed template encoder and a static template bank inherited from Stage 1.

\paragraph{Stage 2 EMA Shallow, Mid, Deep.}
EMA stabilized Stage 2 variants with one, three, or six trainable template encoder layers on the template side, respectively.

\paragraph{Snapshot + KLD.}
A Stage-2 variant that uses a snapshot teacher with KLD regularization but no EMA momentum update.

\paragraph{EMA + ALT.}
EMA-stabilized variants with alternating freeze schedules between product-side and template-side updates.

\paragraph{One-Opt variants.}
Variants that train both towers with a single optimizer, with no refresh, one-epoch refresh, or five-epoch refresh of the template bank.

\paragraph{Transfer Frozen-TE FT.}
A transfer-initialized USPTO-50k fine-tuning run starting from a USPTO-Full Stage-2 checkpoint, with the same Stage-2 candidate budget used in the local regime.

\subsection{Optimization and Candidate Budgets}
\label{app:exp_opt}

Stage 1 uses batch size $256$, AdamW, learning rate $10^{-4}$, weight decay $10^{-2}$, mixed precision training, and symmetric InfoNCE. All Stage-2 variants use AdamW, weight decay $10^{-2}$, gradient clipping at $1.0$, temperature $\tau_{\mathrm{temp}}=0.07$, mixed precision training, and FAISS IVF retrieval for hard-negative mining.

For local frozen-template Stage-2 runs, we evaluate candidate set sizes $|\mathcal{C}| \in \{32,64,128\}$ and use $|\mathcal{C}|=64$ as the main reference in the paper. The EMA variants also use the same main candidate size. The product optimizer uses learning rate $10^{-4}$ and the template optimizer uses learning rate $10^{-5}$. Repeated EMA experiments are reported as mean $\pm$ standard deviation over two runs.

At evaluation time, all local and transfer models use the same test-time template retrieval setting: candidate pool $|\mathcal{C}_{\mathrm{eval}}|=128$, FAISS retrieval pool top-$k=256$.

\subsection{Compute Resources and Approximate Computational Cost}
\label{app:compute_cost}

All experiments were run on a single-GPU desktop workstation with an AMD Ryzen 9 5950X CPU (16 cores / 32 threads), 64\,GiB RAM, and one NVIDIA GeForce RTX 3090 GPU with 24\,GiB memory. The software environment used Ubuntu 24.04.4 LTS, Python 3.11, PyTorch 2.7.1+cu128, and CUDA 12.8 through PyTorch.

Table~\ref{tab:compute_cost} reports approximate per-epoch wall-clock times for representative experiments. These numbers are intended as reproducibility guidance rather than hardware independent performance claims; they depend on implementation details, I/O, and the local software stack. In particular, the EMA based variants are slower than Frozen-TE because they require live template-side scoring, EMA updates, and epoch-level FAISS bank refresh. Snapshot + KLD is the slowest local variant among the reported runs.

\begin{table}[H]
\centering\small
\caption{Approximate computational cost on a single RTX 3090 (24\,GiB). Times are measured as wall-clock training time per epoch on the local compute environment used in this work.}
\label{tab:compute_cost}
\renewcommand{\arraystretch}{1.12}
\setlength{\tabcolsep}{5pt}
\begin{tabular}{@{}ll@{}}
\toprule
\textbf{Experiment} & \textbf{Approx.\ training time per epoch} \\
\midrule
USPTO-50k Stage 1 contrastive pretraining & $\sim 0.8$ min \\
USPTO-50k Stage 2 Frozen-TE, $|\mathcal{C}|=64$ & $\sim 0.9$ min \\
USPTO-50k EMA Shallow & $\sim 16$ min \\
USPTO-50k EMA Mid & $\sim 21$ min \\
USPTO-50k EMA Deep & $\sim 35$ min \\
USPTO-50k Snapshot + KLD & $\sim 45$ min \\
USPTO-50k EMA + ALT & $\sim 0.9$ min in frozen epochs, $\sim 21$ min in update epochs \\
USPTO-Full Stage 1 contrastive pretraining & $\sim 9$ min \\
USPTO-Full Stage 2 Frozen-TE & $\sim 9$ min \\
USPTO-Full $\rightarrow$ USPTO-50k transfer fine-tuning & $\sim 0.9$ min \\
\bottomrule
\end{tabular}
\end{table}

At inference time, each query consists of five serial stages: SMILES tokenization,
product-encoder (PE) forward pass, dense cosine-similarity retrieval over the
pre-built template embedding bank, application of the top-50 templates via RDKit
(yielding up to 4 outcomes per template), and canonicalization and deduplication of
the resulting reactant sets.
Table~\ref{tab:infer_latency} reports mean and 95th-percentile wall-clock times per
component, measured over 200 validation queries after 20 warm-up queries on the same
RTX~3090 workstation described above.
The dominant cost on \emph{both} devices is RDKit template application, which is
inherently single-threaded CPU work and therefore device-independent.
The GPU advantage over CPU-only inference ($10.5$ vs.\ $17.2$\,ms end-to-end) stems
entirely from the faster PE forward pass and dense retrieval step.
The template embedding bank (for USPTO-50K) is built once at startup in under 3\,seconds on GPU and
is fully amortized across all subsequent queries.
Peak GPU memory during inference is approximately 1\,GiB. The PE forward pass requires on average $1.07$\,GFLOPs per query at a mean
SMILES length of 170 tokens.

\begin{table}[H]
\centering\small
\caption{Per-query inference latency on a single RTX~3090 (GPU) and on CPU only,
measured over 200 validation queries (top-50 template retrieval).
RDKit template application is always CPU-bound and therefore takes the same time
on both devices.}
\label{tab:infer_latency}
\renewcommand{\arraystretch}{1.12}
\setlength{\tabcolsep}{5pt}
\begin{tabular}{@{}lcccc@{}}
\toprule
 & \multicolumn{2}{c}{\textbf{GPU}} & \multicolumn{2}{c}{\textbf{CPU}} \\
\cmidrule(lr){2-3}\cmidrule(lr){4-5}
\textbf{Component} & Mean (ms) & p95 (ms) & Mean (ms) & p95 (ms) \\
\midrule
Tokenize              & 0.08  & 0.11  & 0.12  & 0.20  \\
PE encode             & 2.75  & 2.79  & 8.75  & 12.93 \\
Dense retrieval + top-$k$ & 0.12  & 0.13  & 1.06  & 1.18  \\
RDKit apply + dedup   & 7.16  & 12.55 & 7.03  & 14.34 \\
\midrule
\textbf{Total end-to-end} & \textbf{10.50} & \textbf{15.95} & \textbf{17.24} & \textbf{26.62} \\
\textbf{Throughput (q/s)} & \multicolumn{2}{c}{\textbf{95.2}} & \multicolumn{2}{c}{\textbf{58.0}} \\
\bottomrule
\end{tabular}
\end{table}

\subsection{Compared Baselines}
\label{app:baseline_descriptions}

\paragraph{GLN.}
A template-based method that scores reaction templates using graph-structured representations \citep{dai2019retrosynthesis}.

\paragraph{LocalRetro.}
A template-based method that emphasizes local reaction-center information and local reactivity for retrosynthesis prediction \citep{chen2021deep}.

\paragraph{MHNreact.}
A retrieval-oriented template model based on Modern Hopfield Networks for few- and zero-shot reaction template prediction \citep{seidl2104modern}. Because it is the closest retrieval-oriented template prior, we report a separate protocol-caveated comparison in Appendix~\ref{app:mhnreact_comparison} rather than mixing it into the main comparison table.

\paragraph{RetroGFN.}
A template-based method that uses generative flow networks to encourage diverse and feasible retrosynthesis predictions \citep{gainski2025diverse}.

\paragraph{GraphRetro.}
A structured retrosynthesis model that decomposes prediction into reaction-center identification and reactant completion \citep{somnath2021learning}.

\paragraph{MEGAN.}
A graph-edit model that represents reactions as sequences of molecular editing actions \citep{sacha2021molecule}.

\paragraph{Retroformer.}
A transformer-based template-free retrosynthesis model operating over reaction sequence representations \citep{wan2022retroformer}.

\paragraph{Chimera.}
An ensemble retrosynthesis model combining diverse inductive biases \citep{maziarz2024chimera}.

\paragraph{RetroSynFlow.}
A discrete flow-matching model for accurate and diverse single-step retrosynthesis \citep{yadav2025retro}.

\section{Extended Results}
\label{app:extended_results}

This appendix collects result tables omitted from the main paper for space reasons, including extended ablations, long-tail analyses, transfer results, and additional reliability diagnostics.

\subsection{Three-Seed Local Stability Check}
\label{app:three_seed_local}

\begin{table}[H]
\centering\small
\caption{Three-seed local stability check for the main local variants on USPTO-50k. Values are mean $\pm$ std over three independent runs.}
\label{tab:three_seed_local}
\renewcommand{\arraystretch}{1.12}
\setlength{\tabcolsep}{4pt}
\begin{tabular}{@{}lccccc@{}}
\toprule
\textbf{Method} & \textbf{Top-1} & \textbf{Top-3} & \textbf{Top-5} & \textbf{Top-10} & \textbf{TRetr@10} \\
\midrule
Frozen-TE & 61.28$\pm$0.12 & 77.93$\pm$0.09 & 82.51$\pm$0.14 & 85.74$\pm$0.06 & 67.85$\pm$0.15 \\
EMA Mid & \textbf{62.44}$\pm$0.09 & \textbf{81.72}$\pm$0.23 & \textbf{85.29}$\pm$0.12 & \textbf{87.71}$\pm$0.16 & 70.23$\pm$0.22 \\
Snapshot + KLD & 61.98$\pm$0.11 & 81.67$\pm$0.17 & 85.19$\pm$0.12 & 87.50$\pm$0.12 & \textbf{70.39}$\pm$0.31 \\
\bottomrule
\end{tabular}
\end{table}

\subsection{One-Optimizer Co-Training Ablation}
\label{app:oneopt}

\begin{table}[H]
\centering\small
\caption{One-optimizer co-training ablation. All variants use a single AdamW optimizer at $\eta=2\times 10^{-4}$ and no EMA stabilization.}
\label{tab:oneopt_appendix}
\renewcommand{\arraystretch}{1.12}
\setlength{\tabcolsep}{2.8pt}
\fontsize{7.5pt}{9pt}\selectfont
\begin{tabular}{@{}lccccc|ccccc@{}}
\toprule
& \multicolumn{5}{c|}{\textbf{Reaction Acc. (\%)}} & \multicolumn{5}{c}{\textbf{TRetr. (\%)}} \\
\cmidrule(lr){2-6}\cmidrule(l){7-11}
\textbf{Variant} & @1 & @3 & @5 & @10 & @20 & @1 & @3 & @5 & @10 & @20 \\
\midrule
One-Opt (no refresh)   & 49.51 & 70.13 & 75.82 & 80.50 & 82.06 & 20.38 & 35.78 & 44.24 & 55.72 & 66.21 \\
One-Opt (1-ep refresh) & 48.23 & 69.75 & 75.26 & 79.56 & 80.86 & 20.22 & 35.12 & 42.94 & 53.41 & 63.84 \\
One-Opt (5-ep refresh) & 49.55 & 71.63 & 76.58 & 80.54 & 81.64 & 21.76 & 36.96 & 44.58 & 55.60 & 65.31 \\
Stage 1 (reference)    & 50.49 & 72.47 & 78.82 & 83.54 & 85.33 & 21.40 & 38.64 & 47.69 & 59.66 & 71.73 \\
\bottomrule
\end{tabular}
\end{table}

\subsection{Frozen-TE Candidate Size Ablation}
\label{app:frozen_te}

\begin{table}[H]
\centering\small
\caption{Candidate-size ablation for local Stage-2 Frozen-TE. Best value per column is in bold.}
\label{tab:frozen_te_appendix}
\renewcommand{\arraystretch}{1.12}
\setlength{\tabcolsep}{2.8pt}
\fontsize{7.5pt}{9pt}\selectfont
\begin{tabular}{@{}lccccc|ccccc@{}}
\toprule
& \multicolumn{5}{c|}{\textbf{Reaction Acc. (\%)}} & \multicolumn{5}{c}{\textbf{TRetr. (\%)}} \\
\cmidrule(lr){2-6}\cmidrule(l){7-11}
\textbf{$|\mathcal{C}|$} & @1 & @3 & @5 & @10 & @20 & @1 & @3 & @5 & @10 & @20 \\
\midrule
32  & 60.50 & 77.62 & 82.38 & 85.37 & 86.51 & 37.28 & 51.71 & 58.44 & 67.29 & 74.77 \\
64  & \textbf{61.42} & \textbf{78.00} & 82.34 & 85.73 & 87.07 & \textbf{38.72} & \textbf{52.09} & 59.02 & 67.87 & 75.60 \\
128 & 60.78 & 77.98 & \textbf{82.98} & \textbf{86.05} & \textbf{87.45} & 37.16 & 51.37 & \textbf{59.04} & \textbf{68.03} & \textbf{76.14} \\
\bottomrule
\end{tabular}
\end{table}

The candidate size \(|\mathcal{C}|\) in Table~\ref{tab:frozen_te_appendix}
refers to the Stage-2 training candidate set size. At evaluation time, all
local and transfer models use the same two-stage retrieval procedure. First,
FAISS returns a coarse retrieval pool of size \(K_{\mathrm{FAISS}}=4096\) from
the precomputed template bank. Second, these 4096 templates are re-scored by
the model and truncated to an evaluation candidate set of size
\(|\mathcal{C}_{\mathrm{eval}}|=2048\). Template application is then performed
only on the top 50 templates from this re-ranked candidate set, with up to 4
reaction outcomes retained per template. The resulting reactant sets are
canonicalized, deduplicated, and used to compute the final reactant-level
metrics.

\subsection{EMA Depth Ablation}
\label{app:ema_depth}

\begin{table}[H]
\centering\small
\caption{Template-encoder depth ablation for local Stage-2 EMA. This auxiliary depth ablation is reported over three independent runs.}
\label{tab:ema_depth_appendix}
\renewcommand{\arraystretch}{1.12}
\setlength{\tabcolsep}{2.8pt}
\fontsize{7pt}{8.5pt}\selectfont
\begin{tabular}{@{}lccccc|ccccc@{}}
\toprule
& \multicolumn{5}{c|}{\textbf{Reaction Acc. (\%), mean$\pm$std}} & \multicolumn{5}{c}{\textbf{TRetr. (\%), mean}} \\
\cmidrule(lr){2-6}\cmidrule(l){7-11}
\textbf{Variant} & @1 & @3 & @5 & @10 & @20 & @1 & @3 & @5 & @10 & @20 \\
\midrule
EMA Shallow ($K{=}1$) & $61.78\pm0.31$& $81.52\pm0.43$& $85.42\pm0.10$& $\textbf{87.94}\pm0.25$& $\textbf{88.90}\pm0.21$& 36.25& 53.91& 60.83& 70.08& 78.39\\
EMA Mid ($K{=}3$)     & $\mathbf{62.49\pm0.04}$ & $81.59\pm0.07$ & $85.32\pm0.16$ & $87.77\pm0.17$ & $88.70\pm0.04$ & \textbf{37.39} & \textbf{54.65} & 61.54 & 70.11 & \textbf{78.66} \\
EMA Deep ($K{=}6$)    & $62.17\pm0.24$& $\textbf{81.91}\pm0.21$& $\textbf{85.60}\pm0.16$& $87.88\pm0.13$& $88.65\pm0.18$& 36.69& 54.81& \textbf{62.11}& \textbf{70.66}& 78.55\\
Frozen-TE reference   & 61.42 & 78.00 & 82.34 & 85.73 & 87.07 & 38.72 & 52.09 & 59.02 & 67.87 & 75.60 \\
\bottomrule
\end{tabular}
\end{table}

\subsection{Alternating Freeze and Snapshot + KLD}
\label{app:alt_kld}

\begin{table}[H]
\centering\small
\caption{Alternating-freeze ablation for EMA Mid with no KLD.}
\label{tab:alt_appendix}
\renewcommand{\arraystretch}{1.12}
\setlength{\tabcolsep}{2.8pt}
\fontsize{7.5pt}{9pt}\selectfont
\begin{tabular}{@{}lccccc|ccccc@{}}
\toprule
& \multicolumn{5}{c|}{\textbf{Reaction Acc. (\%)}} & \multicolumn{5}{c}{\textbf{TRetr. (\%)}} \\
\cmidrule(lr){2-6}\cmidrule(l){7-11}
\textbf{$T_f{:}T_u$} & @1 & @3 & @5 & @10 & @20 & @1 & @3 & @5 & @10 & @20 \\
\midrule
1:1   & 59.98 & 80.30 & 84.44 & 87.19 & 88.49 & 33.47 & 50.83 & 58.22 & 67.15 & 75.80 \\
5:1   & 60.66 & 79.82 & 84.24 & 87.09 & 88.31 & 34.83 & 50.83 & 58.86 & 67.65 & 76.32 \\
10:2  & 60.74 & 79.52 & 84.16 & 86.91 & 87.97 & 35.62 & 51.77 & 58.90 & 67.97 & 76.46 \\
30:5  & 60.60 & 79.50 & 83.88 & 86.67 & 87.87 & 35.54 & 51.65 & 58.92 & 68.11 & 76.30 \\
\bottomrule
\end{tabular}
\end{table}

\begin{table}[H]
\centering\small
\caption{Snapshot + KLD result on USPTO-50k over three independent runs.}
\label{tab:kld_appendix}
\renewcommand{\arraystretch}{1.12}
\setlength{\tabcolsep}{2.8pt}
\fontsize{7pt}{8.5pt}\selectfont
\begin{tabular}{@{}lccccc|ccccc@{}}
\toprule
& \multicolumn{5}{c|}{\textbf{Reaction Acc. (\%), mean$\pm$std}} & \multicolumn{5}{c}{\textbf{TRetr. (\%), mean}} \\
\cmidrule(lr){2-6}\cmidrule(l){7-11}
\textbf{Variant} & @1 & @3 & @5 & @10 & @20 & @1 & @3 & @5 & @10 & @20 \\
\midrule
Snapshot + KLD ($K{=}3$) & $61.98\pm0.11$& $81.67\pm0.17$& $85.19\pm0.12$& $87.50\pm0.12$& $88.29\pm0.19$& 36.81& 54.60& 61.45& 70.39& 78.11\\
\bottomrule
\end{tabular}
\end{table}

\subsection{Transfer and USPTO-Full In-Domain Results}
\label{app:transfer_results}

\begin{table}[H]
\centering\small
\caption{Transfer evaluation on USPTO-50k. S1 and S2 are zero-shot USPTO-Full checkpoints evaluated with the USPTO-Full template library. S2 FT is USPTO-50k fine-tuning from the USPTO-Full Stage-2 checkpoint.}
\label{tab:transfer_uspto50k_appendix}
\renewcommand{\arraystretch}{1.12}
\setlength{\tabcolsep}{2.8pt}
\fontsize{7.5pt}{9pt}\selectfont
\begin{tabular}{@{}lccccc|ccccc@{}}
\toprule
& \multicolumn{5}{c|}{\textbf{Reaction Acc. (\%)}} & \multicolumn{5}{c}{\textbf{TRetr. (\%)}} \\
\cmidrule(lr){2-6}\cmidrule(l){7-11}
\textbf{Transfer setting} & @1 & @3 & @5 & @10 & @20 & @1 & @3 & @5 & @10 & @20 \\
\midrule
S1 zero-shot & 44.20 & 62.82 & 68.69 & 73.07 & 74.27 & 34.91 & 52.55 & 58.58 & 64.42 & 69.25 \\
S2 zero-shot & 48.27 & 65.05 & 70.47 & 73.95 & 74.67 & 40.96 & 55.86 & 60.76 & 65.37 & 69.47 \\
S2 FT        & \textbf{75.36} & \textbf{90.77} & \textbf{93.77} & \textbf{95.94} & \textbf{96.34} & \textbf{74.17} & \textbf{89.23} & \textbf{92.63} & \textbf{95.28} & \textbf{96.64} \\
\bottomrule
\end{tabular}
\end{table}

\begin{table}[H]
\centering\small
\caption{In-domain evaluation on the USPTO-Full test set.}
\label{tab:usptofull_appendix}
\renewcommand{\arraystretch}{1.12}
\setlength{\tabcolsep}{3.2pt}
\fontsize{7.5pt}{9pt}\selectfont
\begin{tabular}{@{}lccccc|ccccc@{}}
\toprule
& \multicolumn{5}{c|}{\textbf{Reaction Acc. (\%)}} & \multicolumn{5}{c}{\textbf{Template Acc. (\%)}} \\
\cmidrule(lr){2-6}\cmidrule(l){7-11}
\textbf{Model} & @1 & @3 & @5 & @10 & @20 & @1 & @3 & @5 & @10 & @20 \\
\midrule
USPTO-Full Stage 1          & 55.62 & 76.73 & 82.93 & 88.81 & 92.45 & 48.20 & 70.04 & 77.28 & 84.61 & 90.02 \\
USPTO-Full Frozen-TE Stage 2 & \textbf{69.27} & \textbf{84.40} & \textbf{88.36} & \textbf{92.16} & \textbf{94.41} & \textbf{64.42} & \textbf{80.30} & \textbf{85.02} & \textbf{89.66} & \textbf{92.90} \\
\bottomrule
\end{tabular}
\end{table}

\subsection{Auxiliary Long-Tail Analysis: Classification vs.\ Retrieval}
\label{app:longtail_appendix}

As additional support for the paper motivation, we compare two scoring paradigms for template-based retrosynthesis on USPTO-50k: a discriminative classifier over the full template vocabulary and a contrastive retrieval model that ranks templates by cosine similarity in a shared embedding space. This analysis is intended to diagnose long-tail behavior rather than to serve as the primary benchmark comparison for ConRetroBert.

The dataset contains 13{,}721 templates in the global vocabulary. Following the auxiliary analysis setup, test reactions are partitioned into three buckets according to the frequency of the ground-truth template in the training split: \emph{head} ($f>5$, $n=2965$), \emph{tail} ($0<f\leq 5$, $n=1028$), and \emph{unseen} ($f=0$, $n=1012$). The classifier uses a linear classification head on top of the ConRetroBert product encoder, while retrieval ranks templates directly by cosine similarity in the learned product-template embedding space.

This analysis uses an open inference template library for the retrieval scorer. Thus, templates with $f=0$ are unseen in the training split but are present as candidate template strings in the library being ranked. This setting tests whether the learned product-template embedding space can score available template strings that were not observed as training labels. It is not a closed-vocabulary classification setting, because the classifier has no output unit for such labels.

Table~\ref{tab:longtail_appendix} reports exact top-$k$ template accuracy by frequency bucket. The main result is that classification performs best on head templates, while retrieval is substantially stronger on tail templates and can assign nonzero similarity to training-unseen templates available in the inference library. Classification is necessarily zero on the unseen bucket because those labels are outside its output vocabulary.

\begin{table}[H]
\centering\small
\caption{Auxiliary long-tail analysis: template top-$k$ accuracy by ground-truth template frequency bucket.}
\label{tab:longtail_appendix}
\renewcommand{\arraystretch}{1.12}
\setlength{\tabcolsep}{4.0pt}
\begin{tabular}{@{}llccc@{}}
\toprule
\textbf{$k$} & \textbf{Model} & \textbf{Head ($f>5$)} & \textbf{Tail ($0<f\leq 5$)} & \textbf{Unseen ($f=0$)} \\
\midrule
\multirow{2}{*}{1}
& Classifier & \textbf{53.76} & 8.46 & 0.00 \\
& Retrieval  & 53.56 & \textbf{18.68} & \textbf{9.19} \\
\midrule
\multirow{2}{*}{5}
& Classifier & \textbf{85.67} & 33.56 & 0.00 \\
& Retrieval  & 78.31 & \textbf{46.89} & \textbf{26.78} \\
\midrule
\multirow{2}{*}{10}
& Classifier & \textbf{91.84} & 48.05 & 0.00 \\
& Retrieval  & 84.25 & \textbf{61.09} & \textbf{37.25} \\
\midrule
\multirow{2}{*}{50}
& Classifier & \textbf{97.98} & 79.86 & 0.00 \\
& Retrieval  & 93.59 & \textbf{84.63} & \textbf{67.69} \\
\bottomrule
\end{tabular}
\end{table}

This auxiliary analysis should be interpreted narrowly. It is not a comparison between the full ConRetroBert pipeline and a classification baseline, nor is it a fair closed-vocabulary comparison on unseen classes. Instead, it compares two scoring paradigms built on a shared backbone and shows that dense retrieval can score rare or training-unseen template strings when they are present in the inference library. Its role in the paper is to support the motivation for retrieval-based template scoring in sparse template regimes.

\subsection{Extended Reliability Diagnostics}
\label{app:reliability_extended}

\begin{table}[H]
\centering\small
\caption{Applicability rate and unique reactant set diagnostics for selected local models.}
\label{tab:apprate_uniquers_appendix}
\renewcommand{\arraystretch}{1.12}
\setlength{\tabcolsep}{3.0pt}
\fontsize{7.5pt}{9pt}\selectfont
\begin{tabular}{@{}lccccc|ccccc@{}}
\toprule
& \multicolumn{5}{c|}{\textbf{AppRate (\%)}} & \multicolumn{5}{c}{\textbf{UniqueRS}} \\
\cmidrule(lr){2-6}\cmidrule(l){7-11}
\textbf{Method} & @$1$ & @$3$ & @$5$ & @$10$ & @$20$ & @$1$ & @$3$ & @$5$ & @$10$ & @$20$ \\
\midrule
Stage 1          & 44.14 & 40.17 & 36.87 & 30.49 & 21.92 & 0.48 & 1.22 & 1.82 & 2.91 & 4.01 \\
S2 Frozen-TE     & 63.18 & 52.82 & 46.50 & 35.14 & 23.55 & 0.69 & 1.60 & 2.28 & 3.33 & 4.31 \\
S2 EMA Mid       & 59.69 & 49.89 & 43.91 & 34.14 & 23.82 & 0.65 & 1.51 & 2.15 & 3.21 & 4.31 \\
Snapshot + KLD   & 60.90 & 49.30 & 42.18 & 31.69 & 22.07 & 0.66 & 1.49 & 2.06 & 2.96 & 3.98 \\
EMA + ALT (10:2) & 58.32 & 50.04 & 45.58 & 37.41 & 24.99 & 0.63 & 1.53 & 2.26 & 3.57 & 4.54 \\
\bottomrule
\end{tabular}
\end{table}

The diagnostics in Table~\ref{tab:apprate_uniquers_appendix} and
Table~\ref{tab:yield_appendix} measure different stages of the evaluation
pipeline. AppRate@k measures whether retrieved template slots produce any
RDKit output, YieldCov@k measures whether one of the top-k retrieved template
slots generates the ground-truth reactant set, and Reaction Acc@k is computed
only after template application, removal of non-applicable templates,
canonicalization, deduplication, and final reactant ranking. Therefore,
YieldCov@1 is not expected to match Reaction Acc@1. If higher-ranked templates
fail to apply, a lower-ranked template that generates the ground-truth reactant
set can still become the top-ranked final reactant prediction.

\begin{table}[H]
\centering\small
\caption{Yield coverage and yield rate diagnostics for selected local models.}
\label{tab:yield_appendix}
\renewcommand{\arraystretch}{1.12}
\setlength{\tabcolsep}{3.0pt}
\fontsize{7.5pt}{9pt}\selectfont
\begin{tabular}{@{}lccccc|ccccc@{}}
\toprule
& \multicolumn{5}{c|}{\textbf{YieldCov (\%)}} & \multicolumn{5}{c}{\textbf{YieldRate (\%)}} \\
\cmidrule(lr){2-6}\cmidrule(l){7-11}
\textbf{Method} & @$1$ & @$3$ & @$5$ & @$10$ & @$20$ & @$1$ & @$3$ & @$5$ & @$10$ & @$20$ \\
\midrule
Stage 1          & 24.04 & 42.48 & 52.09 & 63.76 & 74.83 & 24.04 & 16.05 & 12.77 & 8.51 & 5.54 \\
S2 Frozen-TE     & \textbf{44.12} & 58.18 & 64.76 & 73.15 & 80.40 & \textbf{44.12} & 22.50 & 16.36 & 10.28 & 6.20 \\
S2 EMA Mid       & 42.08 & 60.46 & 67.13 & 75.07 & \textbf{82.31} & 42.08 & 22.96 & 16.65 & 10.47 & \textbf{6.40} \\
Snapshot + KLD   & 42.04 & \textbf{60.88} & \textbf{67.29} & \textbf{75.46} & 81.90 & 42.04 & \textbf{23.15} & \textbf{16.67} & \textbf{10.53} & 6.36 \\
EMA + ALT (10:2) & 40.44 & 57.14 & 64.42 & 72.87 & 80.22 & 40.44 & 21.72 & 15.82 & 10.07 & 6.20 \\
\bottomrule
\end{tabular}
\end{table}

\subsection{Secondary Template Success Analysis}
\label{app:secondary_success_analysis}

To better understand the gap between template retrieval accuracy and final reactant accuracy, we analyze \emph{secondary successes}: test cases in which the top predicted template does not match the recorded positive template, but still generates the correct ground-truth reactant set after template application and deduplication. This analysis is performed on the local EMA Mid model and treats the recorded benchmark template only as a labeled positive, not as the unique chemically valid template.

\begin{table}[H]
\centering\small
\caption{Primary and secondary top-1 successes for the local EMA Mid model on USPTO-50k. Primary means the correct top-1 reactant prediction is produced by the recorded positive template. Secondary means the correct top-1 reactant prediction is produced by a different template.}
\label{tab:secondary_success_summary}
\renewcommand{\arraystretch}{1.12}
\setlength{\tabcolsep}{6pt}
\begin{tabular}{@{}lr@{}}
\toprule
\textbf{Outcome} & \textbf{Count} \\
\midrule
Total test samples & 5,005 \\
Correct top-1 predictions & 3,129 \\
Primary successes & 1,790 \\
Secondary successes & 1,339 \\
Failed predictions & 1,876 \\
\midrule
Secondary share among correct top-1 predictions & 42.8\% \\
\bottomrule
\end{tabular}
\end{table}

\begin{table}[H]
\centering\small
\caption{Reaction class outcome breakdown for the local EMA Mid model. The secondary rate is the fraction of correct predictions within each class that are secondary rather than primary successes.}
\label{tab:secondary_by_class}
\renewcommand{\arraystretch}{1.10}
\setlength{\tabcolsep}{4pt}
\begin{tabular}{@{}lrrrrr@{}}
\toprule
\textbf{Reaction class} & \textbf{Primary} & \textbf{Secondary} & \textbf{Failed} & \textbf{Correct} & \textbf{Secondary rate} \\
\midrule
Heteroatom Alkylation and Arylation & 440 & 500 & 575 & 940 & 53.2\% \\
Acylation and Related Processes & 417 & 447 & 325 & 864 & 51.7\% \\
Deprotections & 447 & 24 & 353 & 471 & 5.1\% \\
C-C Bond Formation & 115 & 132 & 320 & 247 & 53.4\% \\
Reductions & 228 & 81 & 153 & 309 & 26.2\% \\
Functional Group Interconversion (FGI) & 56 & 41 & 87 & 97 & 42.3\% \\
Heterocycle Formation & 18 & 40 & 33 & 58 & 69.0\% \\
Oxidations & 23 & 35 & 24 & 58 & 60.3\% \\
Protections & 36 & 27 & 5 & 63 & 42.9\% \\
Functional Group Addition (FGA) & 10 & 12 & 1 & 22 & 54.5\% \\
\bottomrule
\end{tabular}
\end{table}

These statistics show that the template retrieval versus reactant accuracy gap is not just noise from mislabeled benchmark templates. Secondary successes account for 42.8\% of all correct top-1 predictions, and they are concentrated in chemically structured classes rather than appearing as isolated accidents. In particular, high secondary rates in classes such as heterocycle formation, oxidations, and C-C bond formation indicate that multiple explicit templates can often realize the same effective retrosynthetic transformation.

The confidence scores of secondary successes are also comparable to those of primary successes: the mean top-1 template score is 0.3095 for secondary cases versus 0.2977 for primary cases, indicating that secondary successes are not merely low-confidence edge cases.

Consistent with this interpretation, the predicted and ground-truth templates remain in the same Schneider reaction class for all analyzed secondary successes, showing that these cases preserve the coarse reaction semantics even when the exact template identifier differs.

\subsection{Qualitative Case Studies of Template Interpretability}
\label{app:case_studies}

\begin{figure*}[ht]
    \centering
    \includegraphics[width=\textwidth]{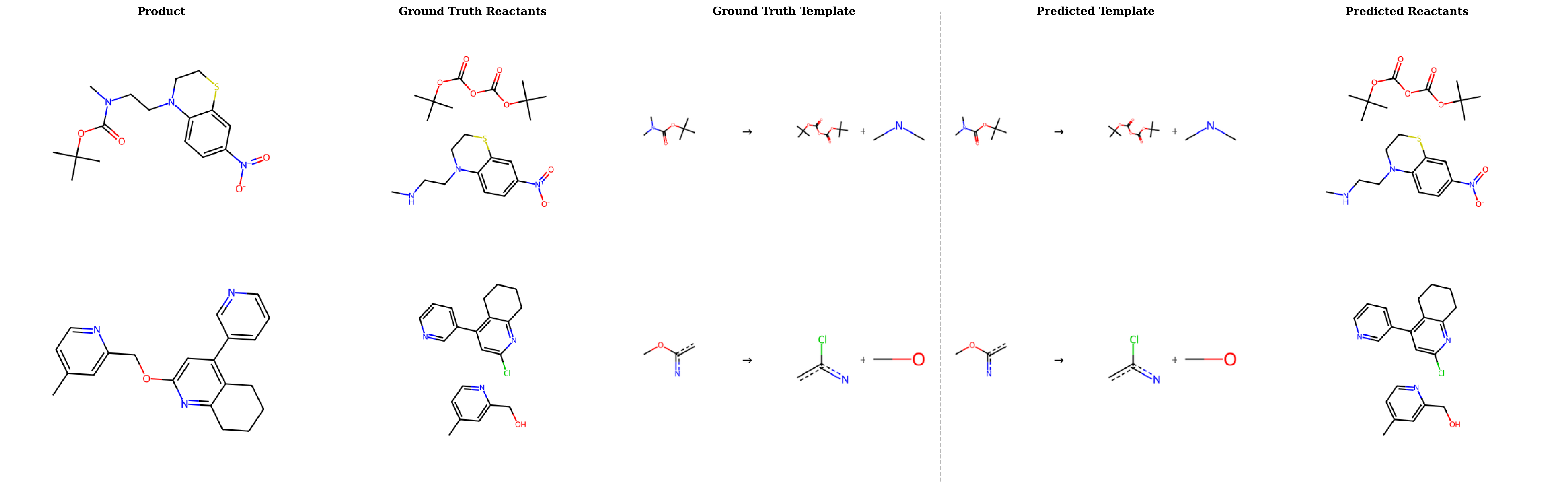}
    \caption{
    \textbf{Exact-template successes.}
    Two representative USPTO-50k test cases in which the top predicted template matches the recorded ground-truth template and yields the correct reactant set. These examples illustrate the most direct form of interpretability in template-based retrosynthesis: the model's decision can be traced to the same explicit transformation rule recorded in the benchmark.
    }
    \label{fig:case_primary_success}
\end{figure*}

\begin{figure*}[ht]
    \centering
    \includegraphics[width=\textwidth]{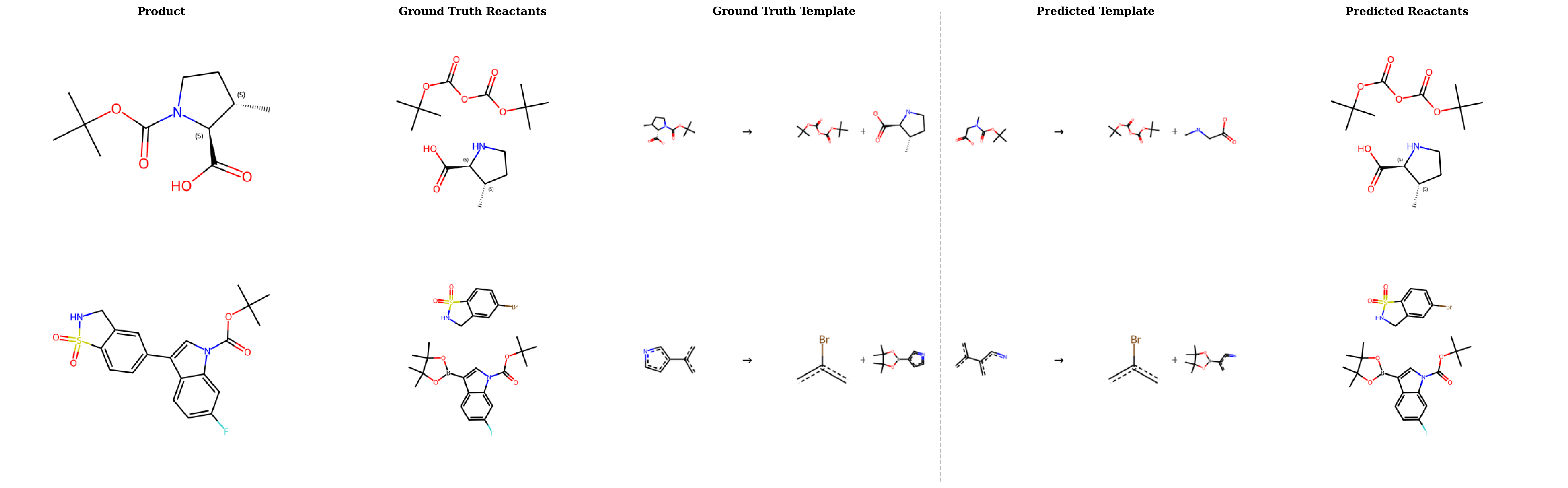}
    \caption{
    \textbf{Alternative-template successes.}
    Two representative USPTO-50k test cases in which the top predicted template differs from the recorded ground-truth template but still generates the correct ground-truth reactant set. These cases help explain the gap between template retrieval accuracy and final reactant accuracy: explicit alternative templates can lead to the correct deduplicated reactant prediction even when the labeled benchmark template is not retrieved first.
    }
    \label{fig:case_secondary_success}
\end{figure*}

\begin{figure*}[ht]
    \centering
    \includegraphics[width=\textwidth]{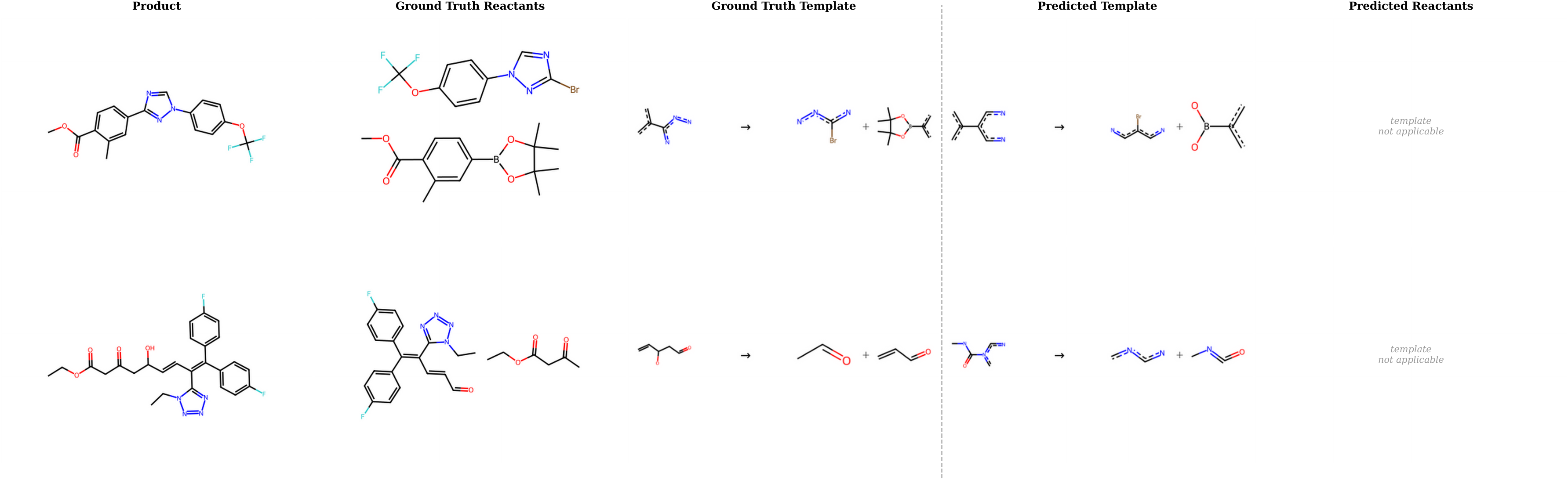}
    \caption{
    \textbf{Failure cases.}
    Two representative USPTO-50k test cases in which the predicted template does not produce the correct reactant set, including cases where the retrieved template is not applicable. Even in failure, the retrieved transformation rule remains explicit and therefore diagnostically useful: the error can be inspected at the level of the proposed reaction rule rather than only at the level of a final incorrect reactant string. Page 1 of the uploaded failure-case figure shows both examples and explicitly marks the predicted template as ``template not applicable.''  
    }
    \label{fig:case_failures}
\end{figure*}

To make the interpretability claim concrete, we show six representative USPTO-50k test cases spanning three qualitatively different outcomes: exact-template successes (Fig. \ref{fig:case_primary_success}), alternative-template successes (Fig. \ref{fig:case_secondary_success}), and failures (Fig. \ref{fig:case_failures}). Each case displays the product, ground-truth reactants, the recorded ground-truth template, the model's predicted template, and the resulting predicted reactants when applicable. Together, these examples illustrate that explicit templates provide a chemically inspectable account of both correct predictions and model errors.

These qualitative cases reinforce three points made elsewhere in the paper. First, when the model succeeds through the recorded benchmark template, the decision is directly interpretable as an explicit reaction rule. Second, correct reactant prediction does not require retrieving the recorded labeled template itself; alternative explicit templates can produce the same ground-truth reactant set after application and deduplication. Third, even failures remain chemically diagnosable because the model exposes the proposed transformation rule, making it possible to distinguish wrong-but-applicable templates from templates that are not applicable to the product at all.

\section{Related Work}
\label{app:related_work_full}
This appendix provides the full related work discussion omitted from the main paper for space.

\paragraph{Single step retrosynthesis.}
Learning based single step retrosynthesis methods are commonly grouped into template-based, semi template, and template free approaches. template-based methods predict an explicit reaction rule and apply it to the product molecule \citep{segler2017neural,coley2017computer,dai2019retrosynthesis}. Semi template methods loosen this dependence on a fixed rule library by predicting reaction centers, graph edits, synthons, or leaving groups before reactant completion \citep{somnath2021learning,sacha2021molecule,zhong2023retrosynthesis,chen2023g,zhao2025single}. Template free methods generate reactants directly as strings, graphs, or latent objects \citep{zheng2019predicting,tu2022permutation,wan2022retroformer,yadav2025retro}. ConRetroBert remains template-based, but revisits this family through retrieval learning rather than global template classification.

\paragraph{Template-based retrosynthesis.}
template-based retrosynthesis is attractive because every prediction is tied to an explicit chemical transformation, making the output easier to inspect, validate, and use inside synthesis planning systems \citep{segler2017neural,coley2017computer,torren2024models}. Classical template-based systems use reaction rules extracted from known reactions and apply them to target molecules through a reaction engine. RetroSim introduced a molecular similarity based approach for ranking reaction templates \citep{coley2017computer}, while neural symbolic retrosynthesis framed template prediction as a learned rule selection problem \citep{segler2017neural}. GLN improved template scoring using graph based representations and remains an important template-based benchmark \citep{dai2019retrosynthesis}. More recent systems have tried to overcome the long tail and rule coverage issues of template-based prediction. LocalRetro emphasizes local reaction center information \citep{chen2021deep}, MHNreact uses associative memory style product template embeddings for few and zero shot generalization \citep{seidl2104modern}, and local template retrieval uses nonparametric neighborhood evidence to support template ranking \citep{xie2023retrosynthesis}. These works suggest that template-based methods are limited not only by template coverage, but also by how products and templates are represented, retrieved, and ranked. ConRetroBert follows this retrieval oriented direction, but trains the product template space with a contrastive objective and then refines it with hard negative listwise ranking.

\subsection{Protocol-Caveated Comparison to MHNreact}
\label{app:mhnreact_comparison}

MHNreact is the closest retrieval-oriented template prior to ConRetroBert, since it frames template prediction through associative memory over reaction templates. We report its published USPTO-50k reactant top-\(k\) accuracy in Table~\ref{tab:mhnreact_appendix} for context. This comparison should be interpreted cautiously: MHNreact reports results under its own USPTO-50k preprocessing and evaluation setup, with standard deviations over five reruns, whereas our local ConRetroBert results use the GLN split, our template application and reactant-set ranking protocol, and three-run means for the main local variants. We therefore keep MHNreact out of the main Table~\ref{tab:sota_local}, which uses prior numbers from the RetroSynFlow comparison table, and report it separately here as a conceptually closest retrieval-based reference.

\begin{table}[H]
\centering\small
\caption{Protocol-caveated comparison to MHNreact on USPTO-50k reactant top-\(k\) accuracy. MHNreact values are the published five-run mean \(\pm\) standard deviation from Seidl et al.; ConRetroBert values are from our local USPTO-50k regime.}
\label{tab:mhnreact_appendix}
\renewcommand{\arraystretch}{1.12}
\setlength{\tabcolsep}{4.5pt}
\begin{tabular}{@{}lcccc@{}}
\toprule
\textbf{Method} & \textbf{Top-1} & \textbf{Top-3} & \textbf{Top-5} & \textbf{Top-10} \\
\midrule
MHNreact~\citep{seidl2104modern} & \(50.5{\pm}0.3\) & \(73.9{\pm}0.3\) & \(81.0{\pm}0.1\) & \(87.9{\pm}0.2\) \\
ConRetroBert Stage 2 Frozen-TE & \(61.28{\pm}0.12\) & \(77.93{\pm}0.09\) & \(82.51{\pm}0.14\) & \(85.74{\pm}0.06\) \\
ConRetroBert Stage 2 EMA Mid & \(62.44{\pm}0.09\) & \(81.72{\pm}0.23\) & \(85.29{\pm}0.12\) & \(87.71{\pm}0.16\) \\
\bottomrule
\end{tabular}
\end{table}

\paragraph{Template free and structured alternatives.}
Template free models have become a dominant alternative because they avoid explicit dependence on a finite template library. Early sequence models such as SCROP treat retrosynthesis as a translation task over reaction SMILES \citep{zheng2019predicting}. Graph2SMILES introduces permutation invariant graph to sequence modeling \citep{tu2022permutation}, and Retroformer pushes transformer based retrosynthesis with reaction aware attention \citep{wan2022retroformer}. Recent generative approaches further improve accuracy, feasibility, or diversity. GFlowNet based retrosynthesis focuses on diverse and feasible reactant generation \citep{gainski2025diverse}, RetroMoE uses a mixture of experts latent translation framework \citep{li2025retromoe}, and RetroSynFlow formulates single step retrosynthesis through discrete flow matching \citep{yadav2025retro}. In parallel, semi template and graph edit methods aim to preserve some structured chemical reasoning while avoiding a strict dependence on a full template library. GraphRetro decomposes retrosynthesis into graph transformation decisions and leaving group completion \citep{somnath2021learning}. MEGAN models reactions as sequences of graph edits \citep{sacha2021molecule}, while Graph2Edits uses an end to end graph generative editing architecture \citep{zhong2023retrosynthesis}. G$^2$Retro predicts synthons and then completes them into reactants \citep{chen2023g}. Retro MTGR jointly models reaction center deduction and leaving group identification through multitask graph representation learning \citep{zhao2025single}. This trend reflects a common tradeoff in the literature: more flexible generation can improve coverage, but often weakens the explicit rule level traceability that template-based systems provide.

\paragraph{Interpretability, reliability, and planning.}
Recent work increasingly emphasizes that retrosynthesis models should be evaluated beyond exact match accuracy. RetroExplainer argues for more interpretable action level reasoning in deep retrosynthesis models \citep{wang2023retrosynthesis}, and Retro MTGR highlights the weak interpretability of many template free systems \citep{zhao2025single}. From the planning side, Models Matter shows that the choice of single step model strongly affects multistep synthesis planning \citep{torren2024models}. Syntheseus further shows that conclusions about retrosynthesis models can change under standardized evaluation settings \citep{maziarz2025re}. These findings motivate careful evaluation and support our emphasis on explicit template predictions as planner compatible actions.

\paragraph{Dense retrieval and hard-negative mining.}
ConRetroBert is also related to dense retrieval methods developed in NLP. Dense Passage Retrieval introduced a dual encoder architecture for efficient retrieval through inner product search \citep{karpukhin2020dense}. ANCE showed that hard negatives mined from a refreshed index can improve dense retrieval training over random negatives \citep{xiong2020approximate}. REALM, Atlas, and EMDR$^2$ further studied retrieval models with periodically refreshed indexes and joint retriever training \citep{guu2020retrieval,izacard2023atlas,singh2021end}. ConRetroBert adapts this retrieval perspective to template-based retrosynthesis: products are queries, templates are retrievable rules, and hard negatives are chemically similar templates that the model must learn to separate from positives.

\paragraph{EMA stabilized encoders.}
EMA encoders have been widely used to stabilize representation learning. MoCo uses a momentum encoder to provide a slowly changing dictionary for contrastive learning \citep{he2020momentum}, and BYOL uses an EMA target network to stabilize self supervised training without negative pairs \citep{grill2020bootstrap}. ConRetroBert uses EMA for a different purpose. The EMA template encoder is not merely a target network; it defines the template bank used for hard-negative mining. This allows the live template encoder to adapt to the Stage 2 ranking loss while the retrieval index evolves smoothly enough to remain useful within each epoch.

\paragraph{Position of our work.}
ConRetroBert is closest to template-based retrieval methods such as GLN, MHNreact, and local template retrieval \citep{dai2019retrosynthesis,seidl2104modern,xie2023retrosynthesis}. It differs in two main ways. First, it separates representation learning from policy learning through contrastive product template pretraining followed by multi positive listwise ranking over hard negative candidate sets. Second, it makes the template encoder trainable during Stage 2 while stabilizing the retrieval index with an EMA shadow encoder. This allows controlled template side adaptation without abandoning the explicit reaction rule interface that makes template-based retrosynthesis useful for interpretation and planning.

\section{Broader Impacts and Dual Use}
\label{app:broader_impacts}

ConRetroBert is a method for template-based single-step retrosynthesis. Its intended beneficial use is to support computer-aided synthesis planning by proposing chemically inspectable reaction templates and reactant sets. Such tools may help researchers analyze synthetic routes, compare disconnection choices, and prioritize experimentally plausible synthesis hypotheses.

At the same time, retrosynthesis models are dual-use technologies because improved synthesis planning could lower the barrier to designing routes for harmful or regulated compounds. This work does not introduce autonomous laboratory execution, purchase recommendations, or experimental protocols, and the model outputs remain template-based predictions that require expert chemical review. Nevertheless, deployment in practical synthesis systems should include appropriate access controls, screening of target molecules and predicted routes against relevant safety and regulatory lists, human expert oversight, and logging or auditing of high-risk use cases. We view these safeguards as necessary when retrosynthesis models are integrated into broader synthesis planning or laboratory automation pipelines.


\end{document}